\documentclass[acmsmall,screen]{acmart}
\setcopyright{none}
\renewcommand\footnotetextcopyrightpermission[1]{} 
\usepackage{algorithm,algpseudocode}
\algnewcommand{\Initialize}[1]{%
  \State \textbf{Initialization in design time:}
  \Statex \hspace*{\algorithmicindent}\parbox[t]{.7\linewidth}{\raggedright #1}
}

\algnewcommand{\Initializes}[1]{%
  \State \textbf{Initialization:}
  \Statex \hspace*{\algorithmicindent}\parbox[t]{.5\linewidth}{\raggedright #1}
}
\algnewcommand{\Observe}[1]{%
  \State \textbf{From the \textit{Observe}:}
  \Statex \hspace*{\algorithmicindent}\hspace*{\algorithmicindent}\hspace*{\algorithmicindent}\parbox[t]{.99\linewidth}{\raggedright #1}
}

\algnewcommand{\Act}[1]{%
  \State \textbf{To the \textit{Act}:}
  \Statex \hspace*{\algorithmicindent}\hspace*{\algorithmicindent}\parbox[t]{.99\linewidth}{\raggedright #1}
}

\algnewcommand{\RMO}[1]{%
  \State \textbf{From \textit{Resource Monitoring}:}
  \Statex \hspace*{\algorithmicindent}\hspace*{\algorithmicindent}\parbox[t]{0.5\linewidth}{\raggedright #1}
}

\algnewcommand{\UQ}[1]{%
  \State \textbf{To \textit{Updating Qtable}:}
  \Statex \hspace*{\algorithmicindent}\hspace*{\algorithmicindent}\parbox[t]{0.9\linewidth}{\raggedright #1}
}

\algnewcommand{\UQN}[1]{%
  \State \textbf{To \textit{Updating Q-Network}:}
  \Statex \hspace*{\algorithmicindent}\hspace*{\algorithmicindent}\parbox[t]{0.9\linewidth}{\raggedright #1}
}

\usepackage{enumitem}

\usepackage{tikz}
\usepackage{pgfplots}
\usepackage{pgfplotstable}
\usepackage{tikz}
\usetikzlibrary{backgrounds,automata}
\usepackage{pgfplotstable}
\usepackage{amsmath,amsfonts}
\usepackage{graphicx}
\usepackage{textcomp}
\usepackage{xcolor}
\usepackage{svg}
\usepackage{algorithm} 
\usepackage{algpseudocode} 
\usepackage{siunitx,booktabs} 

\usepackage{color,soul} 
\usepackage{xcolor}
\usepackage{flushend}
\usepackage{graphicx}
\usepackage{pdfpages}
\usepackage{lineno}
\usepackage{booktabs}
\usepackage{pifont}
\usepackage{tabu}
\usepackage{rotating}
\usepackage{array}
\usepackage[acronym,nomain]{glossaries}              
\newglossary[slg]{symbolslist}{syi}{syg}{Symbolslist} 
\makeglossaries                                   
\usepackage{breqn}
\usepackage{booktabs}
\usepackage{soul}
\usepackage{tikz}
\usepackage{textcomp}
\usepackage{graphicx}
\usepackage{colortbl}
\usepackage[prependcaption, textsize=footnotesize]{todonotes}
\newcommand{\xmark}{\textcolor{black}{\ding{55}}}
\newcommand{\cmark}{\textcolor{black}{\ding{51}}}    
\usepackage{multirow}
\usepackage[numbers,super]{}
\usetikzlibrary{patterns}
\usepackage{pgf}
\usepackage{geometry,array,float}
\usepackage{subcaption}

\begin{document}
\title{Online Learning for Orchestration of Inference in Multi-User End-Edge-Cloud Networks}
\author{Sina Shahhosseini}
\affiliation{%
  \institution{University of California, Irvine}
  \country{USA}
}

\author{DongJoo Seo}
\affiliation{%
  \institution{University of California, Irvine}
  \country{USA}
}

\author{Anil Kanduri}
\affiliation{%
  \institution{University of Turku}
  \country{Finland} 
  }

\author{Tianyi Hu}
\affiliation{%
  \institution{University of California, Irvine}
  \country{USA}
}

\author{Sung-Soo Lim}
\affiliation{%
 \institution{Kookmin University}
 \country{South Korea}
}

\author{Bryan Donyanavard}
\affiliation{%
  \institution{San Diego State University}
  \country{USA}
}

\author{Amir M. Rahmani}
\affiliation{%
  \institution{University of California, Irvine}
 \country{USA}
}

\author{Nikil Dutt}
\affiliation{%
  \institution{University of California, Irvine}
\country{USA}
}
\makeatletter
\let\@authorsaddresses\@empty
\makeatother

\begin{abstract}
Deep-learning-based intelligent services have become prevalent in cyber-physical applications including smart cities and health-care. 
Deploying deep-learning-based intelligence near the end-user enhances privacy protection, responsiveness, and reliability. 
Resource-constrained end-devices must be carefully managed in order to meet the latency and energy requirements of computationally-intensive deep learning services. 
Collaborative end-edge-cloud computing for deep learning provides a range of performance and efficiency that can address application requirements through computation offloading.
The decision to offload computation is a communication-computation co-optimization problem that varies with both system parameters (e.g., network condition) and workload characteristics (e.g., inputs).
On the other hand, deep learning model optimization provides another source of tradeoff between latency and model accuracy.
An end-to-end decision-making solution that considers such computation-communication problem is required to synergistically find the optimal offloading policy and model for deep learning services. 
To this end, we propose a reinforcement-learning-based computation offloading solution that learns optimal offloading policy considering deep learning model selection techniques to minimize response time while providing sufficient accuracy.
We demonstrate the effectiveness of our solution for edge devices in an end-edge-cloud system and evaluate with a real-setup implementation using multiple AWS and ARM core configurations. 
Our solution provides $35\%$ speedup in the average response time compared to the state-of-the-art with less than $0.9\%$ accuracy reduction,
demonstrating the promise of our online learning framework for orchestrating DL inference in end-edge-cloud systems.
\end{abstract}
\maketitle
\pagestyle{plain}

\section{Introduction}

Deep-learning (DL) is advancing real-time and interactive user services in domains such as autonomous vehicles, natural language processing, healthcare, and smart cities~\cite{schmidhuber2015deep}. 
Due to user device resource constraints, deep learning kernels are often deployed on cloud infrastructure to meet computational demands~\cite{barbera2013offload}. 
However, unpredictable network constraints including signal strength and delays affect real-time cloud services~\cite{khelifi2018bringing}. 
Edge computing has emerged to complement cloud services, bringing compute capacity closer to the user-end devices \cite{yousefpour2018all}. 
A collaborative end-edge-cloud architecture is essential to provide deep-learning-based services with acceptable latency to user-end devices \cite{mudassar2018edge}.
The edge paradigm increases offloading opportunities for resource-constrained user-end devices.  
Offloading DL services in a 3-tier end-edge-cloud architecture is a complex optimization problem considering: (i) diversity in system parameters including heterogeneous computing resources, network constraints, and application characteristics, and (ii) dynamicity of DL service environment including workload arrival rate, user traffic, and multi-dimensional performance requirements (e.g., application accuracy, response time) \cite{eshratifar2019jointdnn,shahhosseini2019dynamic,shahhosseini2021exploring}. 

Existing offloading strategies for DL tasks are based on the assumptions that (i) all DL tasks have similar compute intensity and require similar communication bandwidth, (ii) offloading improves performance, and (iii) latency is guaranteed with offloaded tasks. 
However, these assumptions do not hold in practice due to
dynamically varying application and network characteristics, where the computation-communication and accuracy-performance tradeoffs are inconsistent and nontrivial \cite{eshratifar2019jointdnn,shahhosseini2019dynamic,teerapittayanon2017distributed}. 
Under varying system dynamics, such offloading strategies limit the gains from using the edge and cloud resources.  
Further, model optimization techniques such as quantization and pruning can reduce the computation complexity of DL tasks by sacrificing the model accuracy \cite{cheng2017survey,sze2017efficient}. 
Considering model optimization techniques in conjunction with offloading provides opportunities to influence the computation-communication trade-off \cite{taylor2018adaptive}. 
This exposes an alternative to offloading in resource constrained devices executing DL inference. 
Finding the optimal choice between offloading the DL tasks to edge and cloud layers and using optimized models for inference at local devices results in a high-dimensional optimization problem. 


Understanding the underlying system dynamics and intricacies among computation, communication, accuracy, and latency is necessary to orchestrate the DL services on multi-level edge architectures. 
Reinforcement learning is an effective approach to develop such an understanding and interpret the varying dynamics of such systems~\cite{mousavi2016deep,park2019wireless}. 
Reinforcement learning allows a system to identify complex dynamics between influential system parameters and make a decision online to optimize objectives such as response time \cite{sutton2018reinforcement}. 
We propose to employ online reinforcement learning to orchestrate DL services for multi-users over the end-edge-cloud system. 
Our contributions are:
\begin{itemize}
    \item Runtime orchestration scheme for DL inference services on multi-user end-edge-cloud networks. The orchestrator uses reinforcement learning to perform platform offloading and DL model selection at runtime to optimize response time provided accuracy requirements.
    \item Implementation of our online learning solution on a real end-edge-cloud test-bed and demonstration of its effectiveness in comparison with state-of-the-art~\cite{sen2019machine} edge orchestration strategies.
\end{itemize}

\section{Background} 
In this section, we present the relevant background and significance of orchestrating DL workloads on end-edge-cloud architecture.

\subsection{Offloading DL Workloads in End-Edge-Cloud Architecture}
Computation offloading techniques offload an application (or a task within an application) to an external device such as cloud servers \cite{mach2017mobile}. 
Offloading is typically done in order to improve performance or efficiency of devices \cite{barbera2013offload}. 
DL workloads on end-devices are conventionally offloaded to cloud servers, but delay-sensitive services for distributed systems rely on performing inference at the edge as an alternative \cite{yousefpour2018all}. 
Inference at the edge can provide cloud-like compute capability closer to the user devices, reducing data transmission and network traffic load. 
Edge offloading can provide relatively predictable and reliable performance compared to cloud offloading, as there is less workload and network variance \cite{khelifi2018bringing} \cite{kang2017neurosurgeon}.
In the context of the end-edge-cloud paradigm, computation offloading techniques partition workloads and distribute tasks among multiple layers (local device, edge device, cloud servers) such that the performance and efficiency objectives are met. 

The collaborative end-edge-cloud architecture provides execution choices such that each workload can be executed on the device, on the edge, on the cloud, or a combination of these layers. 
Each execution choice effects the performance and energy consumption of the user end device, based on the system parameters such as hardware capabilities, network conditions, and workload characteristics. 
A distributed end-edge-cloud system consists of the following layers:
\begin{itemize}
    \item \textbf{application layer}: provides user level access to a set of services to be delivered by computing nodes
    \item \textbf{platform layer}: provides a set of capabilities to connect, monitor and control end/edge/cloud nodes in a network
    \item \textbf{network layer}: provides connectivity for data and control transfer between different physical devices across multiple
    \item \textbf{hardware layer}: provides hardware capabilities for computing nodes in the system
\end{itemize} 
Each layer presents a diverse set of requirements, constraints, and opportunities to tradeoff performance and efficiency that vary over time. 
For example, the application layer focuses on the user's perception of algorithmic correctness of services, while the platform layer focuses on improving system parameters such as energy drain and data volume migrated across nodes. 
Both application and platform layers have different measurable metrics and controllable parameters to expose different opportunities that can be exploited for meeting overall objectives. 
In the case of DL inference, different DL model structures present opportunities in the application layer, and different computation offloading decisions in a collaborative end-edge-cloud system present opportunities in the platform layer, both for optimizing the execution while meeting required model accuracy. 

\subsection{Intelligence for Orchestration}
Runtime system dynamics affect orchestration strategies significantly in addition to requirements and opportunities. 
Sources of runtime variation across the system stack include workload of a specific computing node, connectivity and signal strength of the network, mobility and interaction of a given user, etc. 
Considering cross-layer requirements, opportunities, and runtime variations provide necessary feedback to make appropriate choices on system configurations such offloading policies.
Identifying optimal orchestration considering the cross-layer opportunities and requirements in the face of varying system dynamics is a challenging problem. 
Making the optimal orchestration choice considering these varying dynamics is an NP-hard problem, while brute force search of a large configuration space is impractical for real-time applications. 
Understanding the requirements at each level of the system stack and translating them into measurable metrics enables appropriate orchestration decision making. 
Heuristic, rule-based, and closed-loop feedback control solutions are not efficient until reaching convergence, which requires long periods of time~\cite{sutton2018reinforcement}. 
To address these limitations, reinforcement learning approaches have been adapted for the computation offloading problem~\cite{sen2019machine}. Reinforcement learning builds specific models based on data collected over initial epochs, and dramatically improves the prediction accuracy~\cite{sutton2018reinforcement}.

\section{Motivation}

 \begin{figure}[tb!]
 \includegraphics[width=1\linewidth]{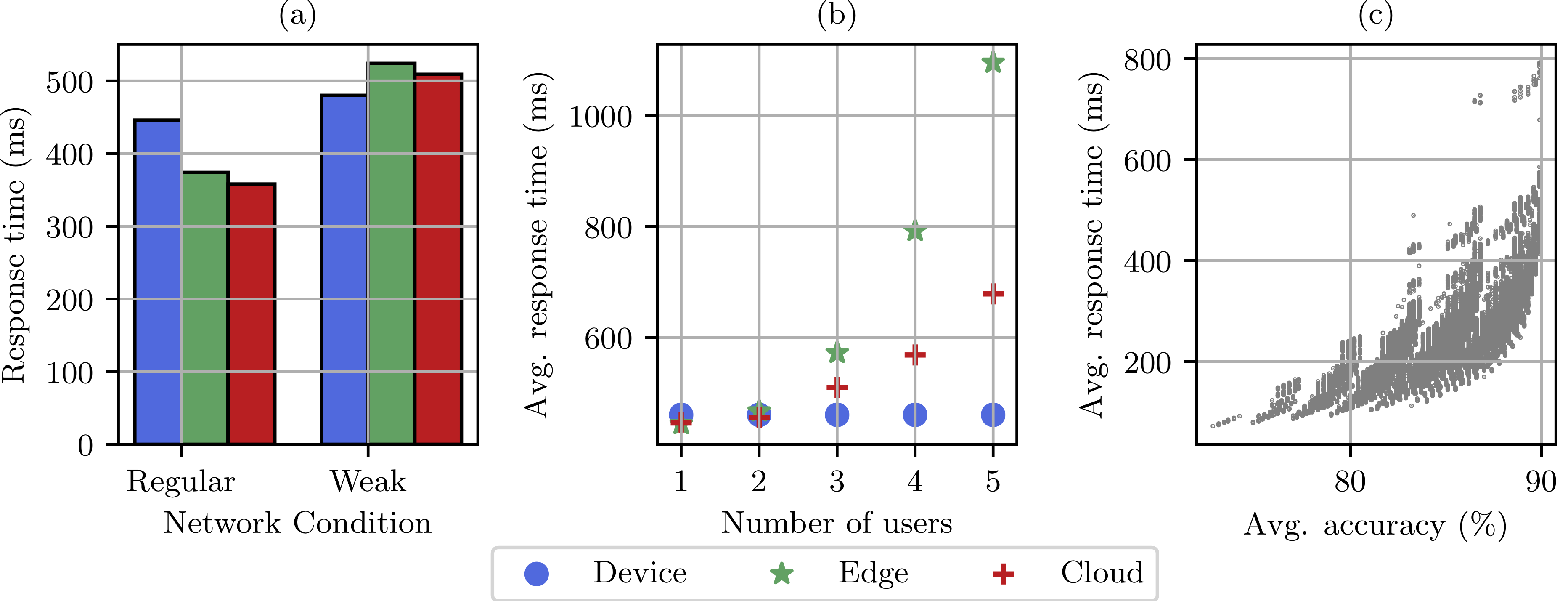}
     \caption {Impact of varying system and application dynamics on performance for MobileNet application. (a) Response time on user-end device, edge and cloud layers with regular and weak network conditions. (b) Average response time with varying number of active users for different computing schemes. (c) Average response time achieved with varying levels of average accuracy.}
     \label{fig.motivation}
\end{figure}

This section presents a comprehensive investigation of DL inference for multi-users in end-edge-cloud systems. 
We examine the scenario using a real setup including five AWS a1.medium instances with single ARM-core as end-node devices connected to an AWS a1.large instance as edge device and an AWS a1.xlarge instance as cloud node. 
We conduct experiments for DL inferences with the MobileNetV1 model while varying (i) network connection, (ii) number of active users, and (iii) accuracy requirement. 
We consider three possible execution choices: (i) on device, (ii) on edge, and (iii) on cloud. 
The device, edge, and cloud execution choices represent executing the inference completely on the local device, on the edge, and on the cloud respectively. 
The detailed specifications for the end-edge-cloud setup appear in Section 5.3. 
    
\subsection{Impact of System Dynamics on Inference Performance}

\paragraph{Network} We consider two possible levels of network connections: (i) a low-latency (regular) network that has the signal strength for better connectivity, and (ii) a high-latency (weak) network that has a weaker signal with poor connectivity. 
Figure \ref{fig.motivation} (a) shows the response time of MobileNet application on user device, edge, and cloud layers with regular and weak networks. 
With a regular network, the response time is highest for executing the application on the user end device. 
The response time decreases as the computation is offloaded to edge and cloud layers, with the higher computational resources. 
With a weak network, the response time of the edge and cloud layers is higher, as the poor signal strength adds delay. 
The response time of the edge node in this case is higher than the cloud layer, given the lower compute capacity of the edge node. 
Performance of the user end device is independent of the network connection, resulting in lowest response time. 
This demonstrates the spectrum of response times achievable with compute nodes at different layers, under varying network constraints. 
For example, the best execution choice with a regular network is the cloud layer, whereas it is the local execution with a weak network.

\paragraph{Users} We examine user variability by considering multiple simultaneously active users ranging from 1 to 5. 
Figure \ref{fig.motivation} (b) shows the average response time with varying number of users. 
The average response time remains constant when running the application on a user end device, i.e., each user executes the application on their local device. 
When offloaded to the edge layer, the average response time increases significantly as the number of users increase. 
This is attributed to the increased network load with multiple simultaneously active users as well as limited resources at the edge layer to handle several user requests concurrently. 
The average response time also increases when offloaded to the cloud layer as the number of simultaneous users increases. 
However, the response time is lower when compared to the edge layer, since the cloud layer has a larger volume of resources to handle multiple simultaneous user requests.

\paragraph{Accuracy} We demonstrate the impact of varying DL models on performance under different system dynamics. 
We select between eight models with Top-5 accuracy between $\%72.8$ and $\%89.9$, while also considering all three layers for execution, and between 1 and 5 simultaneously active users. 
Figure \ref{fig.motivation} (c) shows the average response time achieved with varying levels of average accuracy over a multi-dimensional space of different execution choice and different number of users. 
Each point in Figure \ref{fig.motivation} (c) represents a unique case of an execution choice (among device, edge, and cloud), number of active users (among 1 to 5), and accuracy level. 
We present the average response time achieved with different levels of accuracy. 
As expected, the response time increases with increase in model accuracy. 
However, we observe tradeoffs among different response times between accuracy and number of active users. 
For instance, it is possible to support multiple users within the response time of servicing a single user, by lowering the model accuracy. 

Considering the three major sources of variations in number of users, network conditions, and model accuracy, finding an optimal choice of execution for end-edge-cloud architectures at runtime is challenging. 
As such architectures scale in the number of users and edge nodes, the accuracy-performance Pareto-space becomes increasingly cumbersome for finding an optimal configuration among the fine-grained choices. 
Brute force and smart search algorithms do not offer practically feasible solutions to orchestrate applications in real-time. 
While machine learning algorithms can identify near-optimal configuration choices, they require exhaustive training, considering continuously varying system dynamics. 
We propose to employ online reinforcement learning to understand the volatility of system dynamics and make near-optimal orchestration decisions in real-time to improve the response time of DL inferencing on end-edge-cloud architectures.
\begin{table}[tb!]
\small
\centering
  \caption{Reinforcement Learning Based Works. \textit{CO} represents the computation offloading technique. \textit{HW} and \textit{APP} represents knobs belong to the hardware and application layer, respectively.}
  \vspace{-2mm}
  \begin{tabular}{ c c c c c}
    \toprule
     \parbox[c]{3cm}{\hrule height 0pt width 0pt \centering \textbf{Related Works}} &
     \parbox[c]{1.8cm}{\hrule height 0pt width 0pt \centering \textbf{Real System Evaluation}} &
     \parbox[c]{3cm}{\hrule height 0pt width 0pt \centering \textbf{Multi-User}}&
     \parbox[c]{2.5cm}{\hrule height 0pt width 0pt \centering \textbf{End-to-End}}&
     \parbox[c]{1cm}{\hrule height 0pt width 0pt \centering \textbf{Actions}}\\
    \midrule
    \parbox[t]{3cm}{\centering \cite{qiao2019online,cheng2019dynamic,chen2018optimized,min2019learning,xu2016online}} & \xmark&\xmark&\xmark& CO\\ 
    \parbox[t]{1cm}{\centering \cite{kim2020autoscale}}& \cmark &\xmark&\xmark& CO,HW\\
    \parbox[t]{3cm}{\centering \cite{ke2019joint,li2018deep,lu2020optimization,wei2018dynamic,alam2019autonomic,chen2018decentralized,sen2019machine}}& \xmark&\cmark&\xmark& CO\\ 
    \parbox[t]{1cm}{\centering Ours}& \cmark &\cmark&\cmark& CO,APP\\
    \bottomrule
  \end{tabular}
  \label{table:related}
\end{table}

\subsection{Related Work}
We categorize research related to optimally deploying DL services at the edge in two ways: (i) work related to deploying DL inference tasks over the end-edge-cloud collaborative architecture, and (ii) work related to adopting reinforcement learning methods to optimally offload tasks.

\paragraph{DL Inference in End-edge-cloud Networks} 
Prior works propose frameworks to decompose DL inference into tasks and perform distributed computations. 
In these works, a DL model can be partitioned vertically or horizontally along the end-edge-cloud architecture. 
Generally, DL models are partitioned according to the compute cost of model layers and required bandwidth for each layer to be distributed among the end-edge-cloud~\cite{ran2018deepdecision,xu2019deepwear,jeong2018ionn,kang2017neurosurgeon}. 
These works find the optimal partition points based on traditional optimization techniques and offer design-time optimal solutions.
Some efforts try to reduce the computation overhead of DL tasks through various model optimization methods such as quantization. 
These methods transform or re-design models to fit them into resource-constrained edge devices with little loss in accuracy~\cite{han2015learning, courbariaux2015binaryconnect, mcdanel2017embedded}. 
AdaDeep~\cite{liu2020adadeep} proposes a Deep Reinforcement Learning method to optimally select from a pool of compressed models according to available resources. 
However, AdaDeep relies only on the model selection technique while our work combines computation offloading and model selection techniques to achieve the optimal response time.

\paragraph{Learning-based Offloading} 
Prior works address the offloading problem to optimize different objectives including latency and energy consumption. 
Most of the works formulate the offloading problem with limited number of influential parameters and adopt online learning techniques with numerical evaluation \cite{alam2019autonomic,xu2016online,chen2018decentralized,qiao2019online,wei2018dynamic,ke2019joint,cheng2019dynamic,chen2018optimized,min2019learning,li2018deep}. 
Lu et. al. \cite{lu2020optimization} propose a Deep Recurrent Q-Learning algorithm based on Long Short Term Memory network to minimize the latency for multi-service nodes in large-scale heterogeneous MEC and multi-dependence in mobile tasks. 
The algorithm is evaluated in iFogSim simulator with Google Cluster Trace. 
\cite{sen2019machine} proposes a Q-Learning based algorithm to minimize energy by considering various parameters in task characteristics and resource availability. 
Young Geun et al. \cite{kim2020autoscale} propose a reinforcement learning based offloading technique for energy efficient deep learning inference in the edge-cloud architecture. 
The work focuses on the learning for heterogeneous systems and lacks a comprehensive solution for multi-users end-edge-cloud systems. Haung et al.~\cite{huang2018machine} uses a supervised learning algorithm for complicated radio situations and communication analysis and prediction to make an optimal actions to obtain a high quality of service (QoS). However, our proposed work employs model-free reinforcement learning algorithm to find the optimal orchestration scheme. There have been other efforts which apply game theory algorithms to address the orchestration problem in the network. Apostolopoulos et al.~\cite{apostolopoulos2020risk} propose a decentralized risk-aware data offloading framework using non-cooperative game formulation. The work uses a model-based game theory algorithm to find the optimal offloading decision which makes difference with our proposed model-free reinforcement learning approach. Model-based algorithm use a predictive internal model of the system to seek outcomes while avoiding the consequence of trial-and-error in real-time. The approach is sensitive to model bias and suffers from model errors between the predicted and actual behavior leading to sub-optimal orchestration decisions. Our Model-free RL technique operates with no assumptions about the system's dynamic or consequences of actions required to learn a policy. 

Some works have been applied traditional optimization techniques to optimize the computation offloading problem~\cite{chen2021recent}. Yuan et al.~\cite{yuan2020profit} model a profit-maximized collaborative computation offloading and resource allocation algorithm to maximize the profit of systems and meet the required response time. In another work, Bi et al.~\cite{bi2020energy} propose a partial computation offloading method to minimize the total energy consumed by mobile devices. The work formulates the problem and optimizes using a novel hybrid meta-heuristic algorithm. Considering systems are unknown with dynamic behavior, the traditional optimization techniques are not applicable for runtime decision-making. 
Table 1 positions our work with respect to state-of-the-art solutions. Our solution uses RL to optimally orchestrate DL inference in multi-user networks considering offloading and DL model selection techniques combined together.

\subsection{Contributions}

    
 \begin{figure}[tb!]
    \centering
    \includegraphics[width=0.85\linewidth]{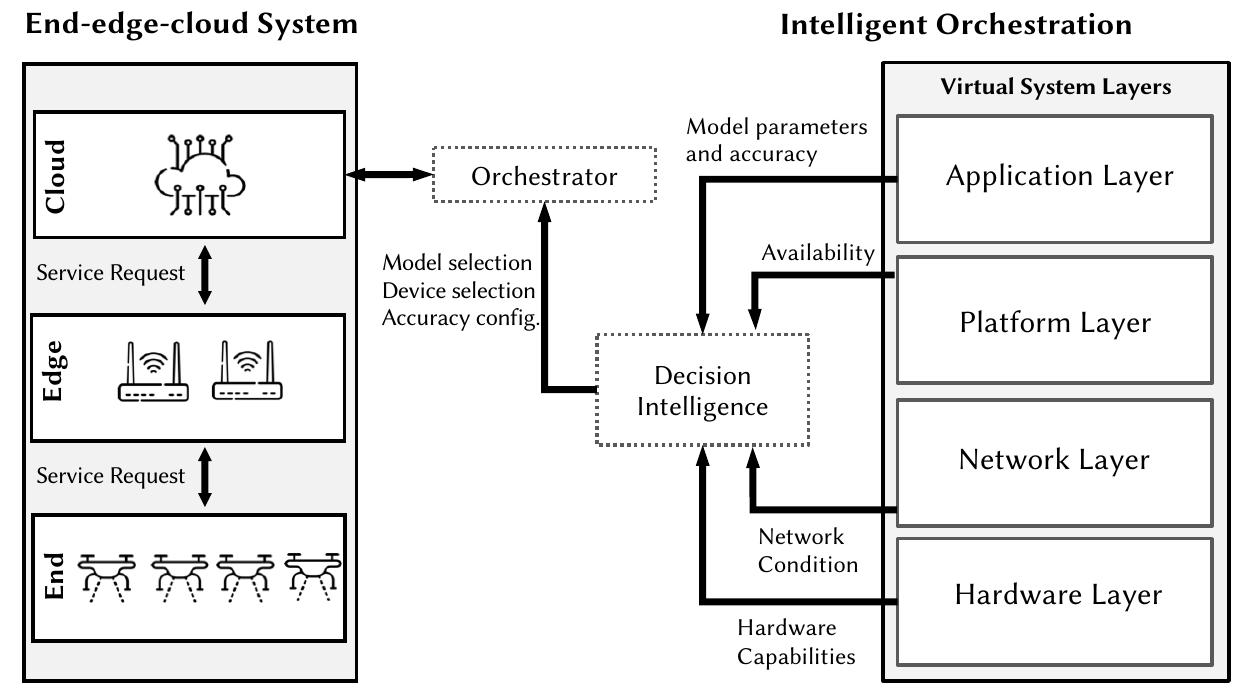}
      \caption{Intelligent orchestration of DL inference in end-edge-cloud architectures.}
      \label{fig.sys_arch}
    \end{figure}

 The ideal DL inference deployment provides maximum inference accuracy and minimum response time. 
 Figure \ref{fig.sys_arch} shows an abstract overview of our target multi-layered architecture for online computation offloading of DL services. We consider three layers viz., user-end device, edge and cloud. Further, we classify this architecture into virtual system layers that include application, platform, network and hardware layers. Each of the virtual system layers provide sensory inputs for monitoring system and application dynamics such as DL model parameters, accuracy requirements, availability of devices for execution, network characteristics, and hardware capabilities. 
 The \textit{Decision Intelligence} component in Figure \ref{fig.sys_arch} periodically monitors resource availability from all virtual system layers to determine appropriate execution choice and DL models to achieve the required QoS (e.g, accuracy, response time). \textit{Decision Intelligence} analyzes the system parameters to make orchestration decisions in terms of model selection, accuracy configuration, and offloading choices. The orchestrator is a software component that is hosted at the cloud layer and enforces the orchestration decisions upon receiving a service request from the user-end devices.
 
 Finding an optimal computation policy including offloading and model selection to optimize objectives (e.g., accuracy, response time) is considered an NP-hard problem. 
 The problem generally can be solved using traditional optimization techniques such as heuristic-based methods, meta-heuristic methods, or exact solutions. 
 Considering systems are unknown with dynamic behavior, the traditional optimization techniques are not applicable for runtime decision-making to optimize objectives.
 Modeling an unexplored high-dimensional system is feasible using model-free reinforcement learning techniques~\cite{sutton2018reinforcement}. Model-free RL operates with no assumptions about the system's dynamic or consequences of actions required to learn a policy. Model-free RL builds the policy model based on data collected through trial-and-error learning over epochs ~\cite{sutton2018reinforcement}. 
 In this work, we use model-free reinforcement learning to deploy DL inference at the edge by considering offloading and model selection.
 Some works have been proposed to address the computation offloading problem using online techniques~\cite{alam2019autonomic,xu2016online,chen2018decentralized,qiao2019online,wei2018dynamic,ke2019joint,cheng2019dynamic,chen2018optimized,min2019learning,kim2020autoscale}. 
 However, there is no relevant work to investigate the integration of online learning with DL inference deployment. 
 Therefore, the literature suffers from some shortcomings that are summarized as follows:
\begin{itemize}
    \item \textbf{\textit{Cross-layer Optimization}:} online solutions have not previously coordinated offloading and model optimizations together. As Table \ref{table:related} shows, all related work relies on only computation offloading (CO). To the best of our knowledge, for the first time, this paper considers both computation offloading and application-level adjustment (APP) together in order to achieve required QoS. 
    \item \textbf{\textit{Real System Evaluation}:} most RL-based solutions in the literature are numerically evaluated. Some works have been proposed and evaluated with simulators. As Table \ref{table:related} shows, the literature lacks a real hardware implementation for online learning framework. This paper implements the online system on real hardware devices which leads to realistic evaluation of online agent's overhead. 
    \item \textbf{\textit{End-to-End Solution}:} end-to-end solution considers a service from the moment a request is issued from the end-node device to delivering results to itself. Table \ref{table:related} illustrates that the literature lacks an end-to-end solution.
\end{itemize}

\begin{table}[t]
\small
\centering
  \setlength\extrarowheight{-3pt}
  \caption{Notation descriptions}
  \begin{tabular}{ c  c}
    \toprule
    \parbox[c]{1cm}{\hrule height 0pt width 0pt \centering \textbf{Notation}}
     & \parbox[c]{7cm}{\hrule height 0pt width 0pt \centering \textbf{Description}} \\
    \midrule
        \parbox[c]{1cm}{\hrule height 0pt width 0pt \centering $S$}         &  \parbox[c]{7cm}{\hrule height 0pt width 0pt  end-node  device}  \\ 

    \parbox[c]{1cm}{\hrule height 0pt width 0pt \centering $E$}         &  \parbox[c]{7cm}{\hrule height 0pt width 0pt  edge device}  \\ 
        \parbox[c]{1cm}{\hrule height 0pt width 0pt \centering $C$}         &  \parbox[c]{7cm}{\hrule height 0pt width 0pt  cloud device}  \\ 
        \parbox[c]{1cm}{\hrule height 0pt width 0pt \centering $P$}         &  \parbox[c]{7cm}{\hrule height 0pt width 0pt   processor  utilization }  \\ 
            \parbox[c]{1cm}{\hrule height 0pt width 0pt \centering $M$}         &  \parbox[c]{7cm}{\hrule height 0pt width 0pt  memory utilization }  \\ 
            \parbox[c]{1cm}{\hrule height 0pt width 0pt \centering $B$}         &  \parbox[c]{7cm}{\hrule height 0pt width 0pt  network condition }  \\
                \parbox[c]{1cm}{\hrule height 0pt width 0pt \centering $o$}         &  \parbox[c]{7cm}{\hrule height 0pt width 0pt offloading decision }  \\
            \parbox[c]{1cm}{\hrule height 0pt width 0pt \centering $o_i^j$}         &  \parbox[c]{7cm}{\hrule height 0pt width 0pt offloading decision for end-node $i$ to resource $j$ }  \\
                \parbox[c]{1cm}{\hrule height 0pt width 0pt \centering $N$}         &  \parbox[c]{7cm}{\hrule height 0pt width 0pt  number of end-node devices }  \\
                \parbox[c]{1cm}{\hrule height 0pt width 0pt \centering $d_k$}         &  \parbox[c]{7cm}{\hrule height 0pt width 0pt  DL model $k$}  \\
        \parbox[c]{1cm}{\hrule height 0pt width 0pt \centering $l$}         &  \parbox[c]{7cm}{\hrule height 0pt width 0pt  number of available DL models }  \\
        \parbox[c]{1cm}{\hrule height 0pt width 0pt \centering $T_{res}^j$}         &  \parbox[c]{7cm}{\hrule height 0pt width 0pt  response time for offloading DL task to resource $j$ }  \\
        \parbox[c]{1cm}{\hrule height 0pt width 0pt \centering $\alpha$}         &  \parbox[c]{7cm}{\hrule height 0pt width 0pt  learning rate }  \\
        \parbox[c]{1cm}{\hrule height 0pt width 0pt \centering $\gamma$}         &  \parbox[c]{7cm}{\hrule height 0pt width 0pt  discount factor }  \\
    \bottomrule
  \end{tabular}
  \label{table:note}
\end{table}
\section{Online Learning Framework}\label{sec.online_framework}
Our goal is to make offloading decisions and inference model selections in order to minimize inference latency while achieving acceptable accuracy.
To do so, we first define the optimization problem, then we propose a reinforcement learning agent to solve the problem. 
Table \ref{table:note} defines the notation used for the problem definition.

\subsection{System Model and Problem Formulation}
All computing devices in the end-edge-cloud system are represented by (S,E,C) where $S=\{S_1,S_2,..,S_n\}$ represents a set of end-node devices whose number is $N$; $E$ represents the edge layer (in our case, a single device); C represents the cloud layer. 
Each end-node device requires a DL inference periodically. 
The inference model is selected from a pool of optimized models where each model has different characteristics including computational complexity and model accuracy. 
All device resources are represented in a tuple $\{P_i,M_i,B_i\}$ where $P_i$ represents processor utilization of device $i$; $M_i$ represents available memory for device $i$; $B_i$ represents network's connection condition between the device $i$ and upper layer's node.

The computation offloading decision determines whether each end-node device should offload an inference to higher-layer computing resources, or perform computation locally. 
The offload decision for each end-node device is represented by a tuple $o_i=\{o^S_i,o^E_i,o^C_i\} $ where $o^j_i$ represents offloading decision to layer $j$. 
If end-node device $i$ executes at layer $j \in \{S,E,C\}$, then $o^j_i=1$; otherwise it must be zero. 
For a given end-node device $i$, the sum of all offloading decisions $\sum_{j}^{\{S,E,C\}} o^j_i$ must equal 1.
$o=\{o_1,o_2,...,o_n\}$ represents the offloading decision vector for all end-node devices. 
The inference model selection determines the implementation of the model deployed for each inference on each end-node device.
Each end-node device $S_i$ can perform inference with one of $l$ DL models $\{d_1,d_2,d_3,...,d_l\}$.

In general, response time is the total time between making a request to a service and receiving the result \cite{ref:analysperf}. 
In our case, response time is the sum of the round trip transmission time from an end-node device to the node that performs the computation, plus the computation time.
Response time $T_{res}$ for a request from end-node device $i$ with offload decision tuple $o_i=\{o_i^S,o^E_i,o^C_i\}$ can be summarized as follows:
\begin{equation}
    T_{res_i}=  o^S_i.T_{res}^{S}+o^E_i.T_{res}^{E}+o^C_i.T_{res}^{C}
\end{equation}
Our objective is to minimize the average response time while satisfying the average accuracy constraint. 
The problem is formulated in the following formula:
\begin{equation}
\begin{aligned}
\textbf{P1:} \min_{}\quad & \frac{1}{N}\sum_{i=1}^{N} T_{res_i}(o_i,d_k)\\
  \textrm{s.t.}\quad & \overline{accuracy}\;>\; threshold
\end{aligned}
\end{equation}
where $\overline{accuracy}$ is the spatial average accuracy for simultaneous DL inferences. 

\subsection{Reinforcement Learning Agent}
 \begin{figure}[tb!]
    \centering
    \includegraphics[width=0.85\linewidth]{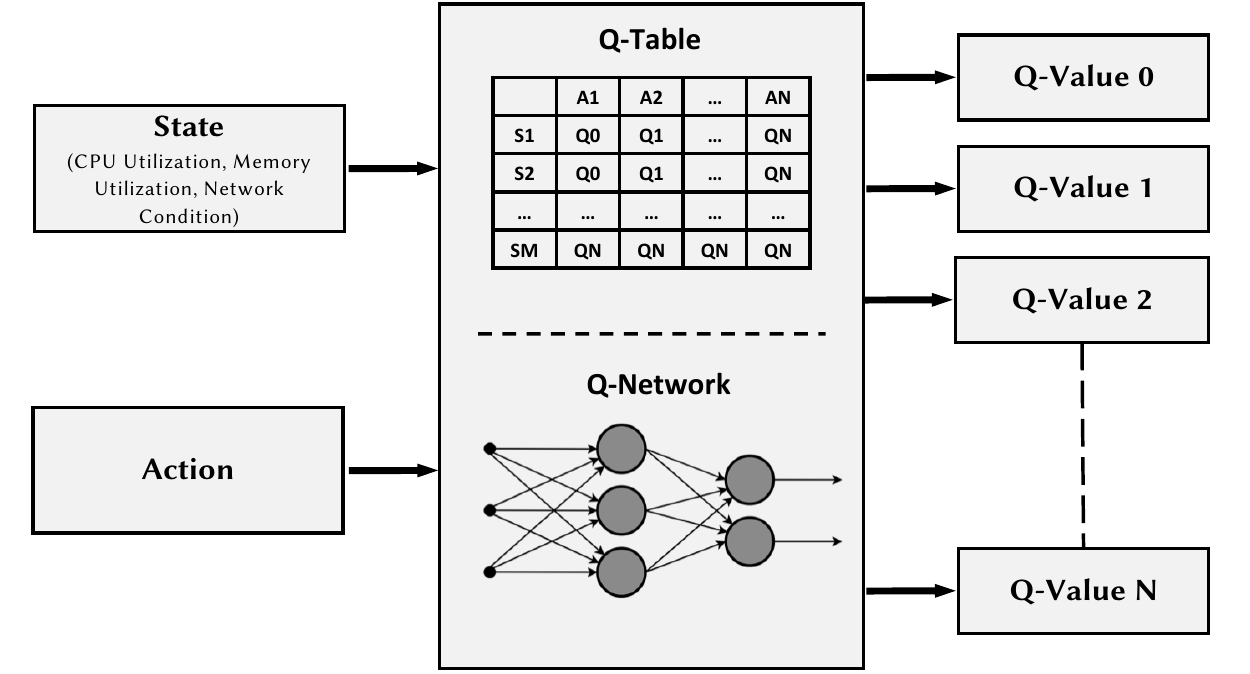}
      \caption{Proposed reinforcement learning agent with Q-Learning and Deep Q-Learning algorithms. Q-Learning uses a Q-Table to store $Q(S,A)$ values, Deep Q-Learning estimates Q-Values with a neural network architecture.}
      \label{fig.agent}
    \end{figure}
Reinforcement learning (RL) is widely used to automate intelligent decision making based on experience. Information collected over time is processed to formulate a policy which is based on a set of rules. Each rule consists three major components viz., (a) state, (b) action, and (c) reward. Among the various RL algorithms \cite{sutton2018reinforcement}, Q-learning has low execution overhead, which makes it a good candidate for runtime invocation. However, it is ineffective for large space problems. There are two main problems  with Q-learning for large space problems \cite{mnih2015human}: (a) required memory to save and update the Q-Values increases as the number of actions and state increases. (b) required time to populate the table with accurate estimates is impractical for the large Q-table. In our case, increasing number of users will increase the problem's space dimension. The reason is more number of users leads to more number of rows and columns in the Q-table. Therefore, it takes more time to explore every state and update the Q-values. Due to the curse of dimensionality, function approximation is more appealing \cite{mnih2015human}. The Deep Q-Learning (DQL) algorithm combines the Q-Learning algorithm with deep neural networks. DQL uses Neural Network architecture to estimate the Q-function by replacing the need for a table to store the Q-values.
In this work, we build an RL agent using two reinforcement learning algorithms: (a) an epsilon-greedy \textbf{Q-Learning} and (b) a \textbf{Deep Q-Learning} algorithms. We evaluate the RL agent with the mentioned algorithms considering different problem complexities. Figure \ref{fig.agent} depicts high-level black diagram for our agent. The RL agent is invoked at runtime for intelligent orchestration decisions. In general, the agent is composed as follows:

\textit{\textbf{State Space:}} 
Our state vector is composed of CPU utilization, available memory, and bandwidth per each computing resource.
Table \ref{table:SD} shows the discrete values for each component of the state. 
The state vector at time step $\tau$ is defined as follows:
\begin{equation}
\begin{split}
S_{\tau}=\{P^{E},M^{E},B^{E},P^{C},M^{C},B^{C},P^{S_1},M^{S_1},B^{S_1},...,
P^{S_n},M^{S_n},B^{S_n}\}
\end{split}
\end{equation}

\textit{\textbf{Action Space:}} The action vector consists of which inference model to deploy, and which layer to assign the inference. 
We limit the edge and cloud devices to always use the high accuracy inference model, and the end-node devices have a choice of $l$ different models. 
Therefore, the action space is defined as $a_{\tau}=\{o^i,d_j\}$ where $i \in \{S,E,C\}$ and $d_j \in \{d_1,d_2,...,d_l\}$. 

\begin{table}[h]
\small
\caption{State Discrete Values}
\centering
  \begin{tabular}{c c c}
    \toprule
     \parbox[c]{0.7cm}{\hrule height 0pt width 0pt \centering \textbf{State}}&
     \parbox[c]{4cm}{\hrule height 0pt width 0pt \centering \textbf{Discrete Values}}&
     \parbox[c]{2.7cm}{\hrule height 0pt width 0pt \centering \textbf{Description}}\\
    \midrule
    \parbox[c]{2.7cm}{\centering $P^{S_i}$}& Available, Busy & End-node CPU Utilization\\
    \parbox[c]{2.7cm}{\centering $M^{S_i}$}& Available, Busy& End-node Memory Utilization\\ 
    \parbox[c]{2.7cm}{\centering $B^{S_i}$}& Regular, Weak & End-node Available Bandwidth\\ 
    \parbox[c]{2.7cm}{\centering $P^{E}$}& Nine discrete levels & Edge CPU Utilization\\
    \parbox[c]{2.7cm}{\centering $M^{E}$}& Available, Busy & Edge Memory Utilization\\
    \parbox[c]{2.7cm}{\centering $B^{E}$}& Regular, Weak & Edge Available Bandwidth\\ 
    \parbox[c]{2.7cm}{\centering $P^{C}$}& Nine discrete levels & Cloud CPU Utilization\\
    \parbox[c]{2.7cm}{\centering $M^{C}$}& Available, Busy& Cloud Memory Utilization\\ 
    \parbox[c]{2.7cm}{\centering $B^{C}$}& Regular, Weak & Cloud Available Bandwidth\\ 
    \bottomrule
  \end{tabular}
  \label{table:SD}
\end{table}

\textit{\textbf{Reward Function:}} The reward function is defined as the negative average response time of DL inference requests.
In our case, the agent seeks to minimize the average response time.
To ensure the agent minimizes the average response time while satisfying the accuracy constraint, the reward $R$ is calculated as follows:
\begin{equation}
\begin{aligned}
\text{if $\overline{accuracy}>$  threshold:}&&\\
&R_{\tau} \gets -Average\;Response\;Time&\\
\text{else:}&&\\
&R_{\tau} \gets -Maximum\;Response\;Time &\\
\end{aligned}
\end{equation}
To apply the accuracy constraint, the minimum possible reward is assigned when the accuracy threshold is violated. On the other hand, when the selected action satisfies the average accuracy constraint, the reward is negative average response time.
\subsubsection{Q-Learning Algorithm}
Q-Learning algorithm is a model-free reinforcement learning algorithm to learn the value of an action in a particular state. The algorithm does not require a model of the environment where it can handle problems with stochastic transitions and rewards without requiring adaptations. The Q-Learning algorithm stores data in a Q-table. The structure of a Q-Learning agent is a table with the states as rows and the actions as the columns. Each cell of the Q-table stores a Q-value, which estimates the cumulative immediate and future reward of the associated state-action pair. Epsilon-greedy is a common enhancement to Q-Learning that helps avoid getting stuck at local optima \cite{sutton2018reinforcement}.
Algorithm \ref{alg:ql} defines our agent's logic with the epsilon-greedy Q-Learning:
\begin{enumerate}[align=left,labelwidth=25pt,leftmargin=40pt]
    \item[\textbf{Line}] \textbf{Description}
	\item[3:] First the agent determines the current system state from the resource monitors.
	\item[4-8:] Next, either the state-action pair $(S_\tau,A_\tau)$ with the highest $Q$-value is identified to choose the next action to take, or a random action is selected with probability $\epsilon$. 
	\item[9-10:] The selected action is applied and normal execution resumes. After all inferences are completed, the reward $R_\tau$ for the execution period is calculated based on measured response time. 
	\item[11-12:] Based on the resource monitors, the new state $A_{\tau+1}$ is identified, along with the state-action pair with highest Q-value.
	\item[13:] The Q-value of the previous state-action pair is updated.
	\item[14:] The current state is updated, and the loop continues.
\end{enumerate}

\begin{algorithm}[!t]
\caption{Q-Learning Algorithm}
\small
\label{alg:ql}
\begin{algorithmic}[1]
    \Initialize{$\tau$ represents time step\\ $S_{\tau}$ represents state at $\tau$ \\ $A_{\tau}$ represents action at $\tau$}
    \While{system is on}
        \RMO { $S_{\tau}$ $\gets$ State at step $\tau$ }
        \If{$RAND$ $<$ $\epsilon$}
            \State Choose random action $A_{\tau}$
        \Else
             \State Choose action $A_{\tau}$ with largest $Q(S_{\tau},A_{\tau})$
        \EndIf
        \State Monitor the response time for each devices
        \State Calculate reward $R_{\tau}$
        \RMO { $S_{\tau+1}$ $\gets$ State at step $\tau+1$ }
        \State Choose action $A_{\tau+1}$ with the largest $Q(S_{\tau+1},A_{\tau+1})$
        \UQ {$Q(S_{\tau},A_{\tau})$ $\gets$ $Q(S_{\tau},A_{\tau})+\alpha[R_{\tau}+\gamma.Q(S_{\tau+1},A_{\tau+1})-Q(S_{\tau},A_{\tau})]$ }
        \State $S_{\tau}$$\gets$ $S_{\tau+1}$
        \label{alg:line:delta}
    \EndWhile
\end{algorithmic}
\end{algorithm}

\subsubsection{Deep Q-Learning Algorithm} 
Q-Learning has been applied to solve many real-world problems. 
However, it is unable to solve high-dimensional problems with many inputs and outputs \cite{mnih2015human} as it is impractical to represent the Q-function as a Q-table for large pair of $S$ and $A$. 
In addition, it is unable to transverse $Q(S,A)$ pairs. 
Therefore, a neural network is used to estimate the Q-values. 
The Deep Q-learning Network (DQN) inputs include current state and possible action, and outputs the corresponding Q-value of the given action input. 
The neural network approximation is capable of handling high dimensional space problems \cite{winder2020reinforcement}. 
One of the main problems with Deep Q-Learning is stability \cite{mnih2015human}. 
In order to reduce the instability caused by training on correlated sequential data, we improve the DQL algorithm with \textit{replay buffer} technique \cite{lin1992self}. 
During the training, we calculate the loss and its gradient using a mini-batch from the buffer. 
Every time the agent takes a \textit{step} (moves to the next state after choosing an action), we push a \textit{record} into the buffer. 
Algorithm 2 defines Deep Q-Learning algorithm which is described below:
\begin{enumerate}[align=left,labelwidth=25pt,leftmargin=40pt]
    \item[\textbf{Line}] \textbf{Description}
	\item[4:] First the agent determines the current system state from the resource monitors.
	\item[4:9] Next, either the state-action pair $S_\tau,A_\tau$ with the highest Q-value estimated by neural network ($\theta$) is identified to choose the next action to take, or a random action is selected with probability $\epsilon$. 
	\item[10:11] The selected action is applied and normal execution resumes. After all inferences are completed, the reward $R_\tau$ for the execution period is calculated based on measured response time.
    \item[12:] At each time step, each record ($S_{\tau}$,$A_{\tau}$,$R_{\tau}$,$S_{\tau+1}$) is added to a circular buffer $D$ called the \textit{replay buffer}.
	\item[13:] We randomly sample \textit{Batch Size} \textit{records} from the buffer and then feed it to the network as mini-batch.
	\item[14:] We calculate the temporal difference loss on the mini-batch and perform a gradient descent calculation to update the network. The \textit{temporal difference loss} function calculates the \textit{mean-square error} of the predicted and target Q-values as the loss of the mini-batch.
	\item[15:] The current state is updated, and the loop continues.
\end{enumerate}

\begin{algorithm}[!b]
\caption{Deep Q-Learning Algorithm with Experience Replay}
\small
\label{alg:dql}
\begin{algorithmic}[1]
    \Initialize{$\tau$ represents time step\\ $S_{\tau}$ represents state at $\tau$ \\ $A_{\tau}$ represents action at $\tau$\\ Initialize replay buffer $D$ to capacity $N$\\ Initialize action-value function $Q$ with random weight $\theta$}
    \For{epoch = 1, Epochs}
        \For{episode = 1, Episodes}
        \RMO { $S_{\tau}$ $\gets$ State at step $\tau$ }
        \If{$RAND$ $<$ $\epsilon$}
            \State Choose random action $A_{\tau}$
        \Else
             \State Choose action $A_{\tau}$ with largest $Q_{\theta}(S_{\tau},A_{\tau})$
        \EndIf
        \State Monitor the response time for each devices
        \State Calculate reward $R_{\tau}$
        \State Store the record ($S_{\tau}$,$A_{\tau}$,$R_{\tau}$,$S_{\tau+1}$) into buffer D
       \State Sample random mini-batch of records from buffer D
        \UQN {Compute temporal difference loss with respect to the network parameter $\theta$}
        \State $S_{\tau}$$\gets$ $S_{\tau+1}$
        \label{alg:line:delta}
   \EndFor 
 \EndFor
\end{algorithmic}
\end{algorithm}

There is few study about the theoretical analysis of the computational complexity of reinforcement learning because of the problem itself that reinforcement learning solves is hard to be explicitly modeled.
Reinforcement learning is in the nature of trail-and-error and exploration-and-exploitation, which involves randomness and makes it difficult to be theoretically analyzed.
 
\noindent The complexity of problem for Bruteforce strategy is discussed in the following: Brute-force strategy searches the entire State$\times$Action space of the problem and sort corresponding Q-values in order to find out the optimal action. Therefore, we can define the state space complexity as follows:
\begin{align}
      (L_{CPU}\times L_{Network} \times L_{Memory})^{N}(L^{\prime}_{CPU}\times L'_{Network} \times L'_{Memory})^{2}  
\end{align}
where, $N$ stands for the number of end-devices. $L_{CPU}$, $L_{Memory}$, and $L_{Network}$ represent the number of CPU, Memory, and Network condition levels for end devices, respectively. In addition, $L^{\prime}_{CPU}$, $L^{\prime}_{Memory}$, and $L^{\prime}_{Network}$ represent the number of CPU, Memory, and Network condition levels for edge and cloud devices, respectively. Besides, the action space is defined as $(Number\;of\; Actions)^N$.
Therefore, the complexity is defined as follows:
\begin{align}
  (L_{CPU}\times L_{Network} \times L_{Memory})^{N}\times(L^{\prime}_{CPU}\times L'_{Network} \times L'_{Memory})^{2} \times (Number\;of\; Actions)^N      
\end{align}

The reinforcement learning agent requires distinct state-action pairs for training the Deep Q-network. To generate distinct state-action pair vectors, our proposed framework supports execution requests that are submitted by all the end-devices synchronously. With synchronous requests, we eliminate the discrepancy of different optimal actions for the same state vector.

\section{Framework Setup}
In this section we describe our proposed framework for dynamic computation offloading based on online learning, targeted at multi-layered end-edge-cloud architecture.

\begin{figure*}[t]
    \centering
    \includegraphics[width=\linewidth]{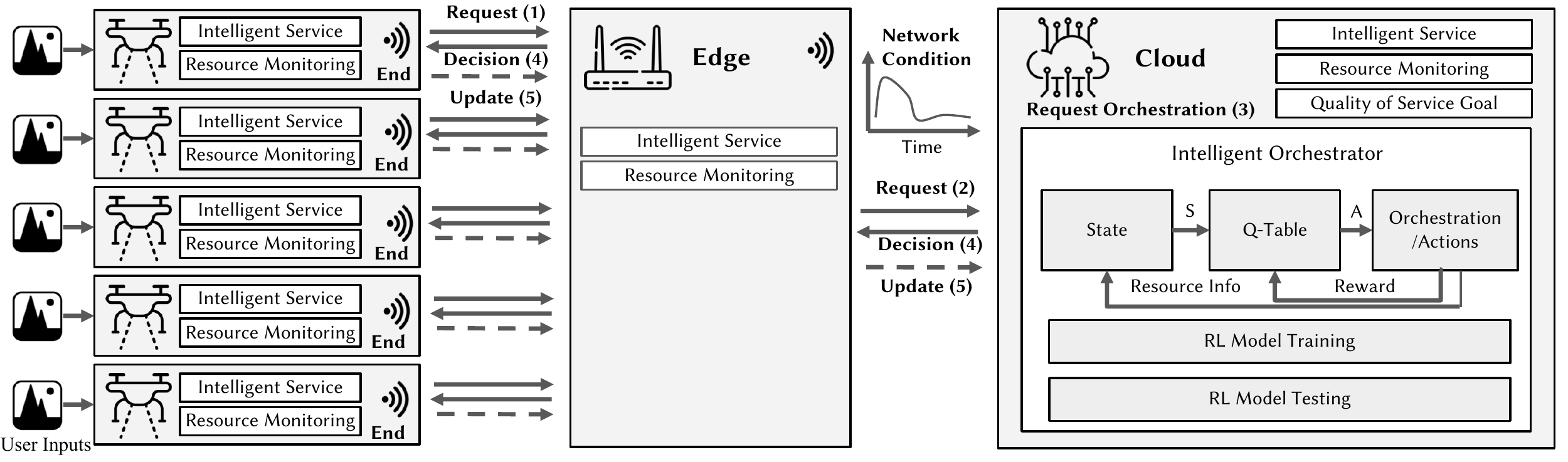}
      \caption{Orchestration framework with online learning for orchestrating DL inference.}
       \label{fig:framework}
\end{figure*}

\subsection{Framework Architecture} 
Figure \ref{fig:framework} shows our proposed framework for end-edge-cloud architecture, integrating service requests, resource monitoring, and intelligent orchestration. The \textit{Intelligent Orchestrator} (IO) acts as an RL-agent for making computation offloading and model selection decisions. 
The end-device layer consists of multiple user-end devices. Each end-device has two software components: (i) \textit{Intelligent Service} - an image classification kernel with DL models of varying compute intensity and prediction accuracy; (ii) \textit{Resource Monitoring} - a periodic service that collects devices' system parameters including CPU and memory utilization, and network condition, and broadcasts the information to the edge and cloud layers.
Both the edge and cloud layers also have the \textit{Intelligent Service} and \textit{Resource Monitoring} components. 
The \textit{Intelligent Orchestrator} acts a centralized RL-agent that is hosted at the cloud layer for inference orchestration. The agent collects resource information (e.g., processor utilization, available memory, available bandwidth) from Resource Monitoring components throughout the network. 
The agent also gathers the reward information (i.e., response time) from the environment in order to learn an optimal policy. 
The agent builds the Q-function based on the RL algorithm. It builds a Q-Table for Q-Learning algorithm and a Q-Network for Deep Q-Learning algorithm based on cumulative reward obtained from the environment over time. 
\textit{Quality of Service Goal} provides the required QoS for the system (i.e., the accuracy constraint).

 Figure~\ref{fig:framework} illustrates the procedure step-wise of the inference service in our framework. 
 The end-device layer consists of resource-constrained devices that periodically make requests to a DL inference service (step 1). 
 The requests are passed through the edge layer (step 2) to the cloud device to be processed by \textit{Intelligent Orchestrator} (step 3). 
 The agent determines where the computation should be executed, and delivers the \textit{Decision} to the network (step 4). 
 Each device updates the agent after it performs an inference with the response time information of the requested service (step 5). 
 In addition, all devices in the framework send the available resource information including the processor utilization, available memory, and network condition to the cloud device (step 5). 
 
 \begin{table}[b]
\small
\caption{MobileNet Models \cite{howard2017mobilenets}}
\centering
  \begin{tabular}{c c c  c c c}
    \toprule
    \parbox[c]{0.5cm}{\hrule height 0pt width 0pt \centering \textbf{\#}}&
     \parbox[c]{2.7cm}{\hrule height 0pt width 0pt \centering \textbf{Model}}&
     \parbox[c]{2cm}{\hrule height 0pt width 0pt \centering \textbf{Million MACs}}&
     \parbox[c]{0.7cm}{\hrule height 0pt width 0pt \centering \textbf{Type}}&
     \parbox[c]{2cm}{\hrule height 0pt width 0pt \centering \textbf{Top-1 Accuracy (\%)}}&
     \parbox[c]{2cm}{\hrule height 0pt width 0pt \centering \textbf{Top-5 Accuracy (\%)}}\\
    \midrule
    $d0$ &\parbox[c]{2.8cm}{\centering 1.0 MobileNetV1-224}& 569& FP32 & 70.9 & 89.9\\   
    $d1$&\parbox[t]{2.8cm}{\centering 0.75 MobileNetV1-224} & 317 & FP32 & 68.4 & 88.2\\
    $d2$&\parbox[t]{2.8cm}{\centering 0.5 MobileNetV1-224}& 150& FP32&63.3&84.9\\ 
    $d3$&\parbox[t]{2.8cm}{\centering 0.25 MobileNetV1-224} & 41& FP32&49.8&74.2\\ 
    $d4$&\parbox[t]{2.8cm}{\centering 1.0 MobileNetV1-224} & 569& Int8&70.1&88.9\\ 
    $d5$&\parbox[t]{2.8cm}{\centering 0.75 MobileNetV1-224}& 317& Int8&66.8&87.0\\ 
    $d6$&\parbox[t]{2.8cm}{\centering 0.5 MobileNetV1-224}&150& Int8&  60.7 &83.2\\
    $d7$&\parbox[t]{2.8cm}{\centering 0.25 MobileNetV1-224} &41& Int8&48.0&72.8\\ 
    \bottomrule
  \end{tabular}
\vspace{1mm}
  \label{table:accuracy}
\end{table}

\subsection{Benchmarks and Scenarios} 
MobileNets are small, low-latency deep learning models trained for efficient execution of image classification on resource-constrained devices~\cite{howard2017mobilenets}. For DL workloads, we consider MobileNetV1 image classification application as the benchmark  \cite{howard2017mobilenets}. We deploy the MobileNetV1 service for end-node classification. We consider eight different MobileNet models ($d0$ through $d7$) with varying levels of accuracy and performance. Each model among $d0$ through $d7$ has varying number of Multiply-Accumulate units (MACs), MAC width and data format (e.g., FP32 and Int8), exposing models with different accuracy-performance trade-offs. Table \ref{table:accuracy} summarizes the MobileNet models we consider, detailing the number of Multiply-Accumulates (MACs), MAC width and data formats (e.g., FP32 and Int8). The multiplier width is used to reduce a network's size uniformly at each layer. For a given layer and multiplier width, the number of input channels and the number of output channels is decreased and increased, respectively, by a factor of the width multiplier. During the orchestration phase, we select an appropriate model from $d0$-$d7$ to achieve the target level of classification accuracy while maximizing the performance.

Our framework supports multiple end-devices, networked with edge and cloud layers. For evaluation purposes, we set the maximum number of simultaneously active user devices to five. Each user-end device is connected to a single edge device, and can request a DL inference service to the cloud layer. The cloud layer hosts the IO that contains the RL agent, which handles the inference service requests. Upon on each service request, the RL agent is invoked to determine: (i) where the request should be processed and (ii) what DL model should be executed for the corresponding request. 
The RL agent's goal is to minimize average response time for all end-node devices while satisfying the accuracy constraint. 
This enforces quality control by imposing a strict threshold on the average DL model accuracy. In this work, we conduct experiments under four unique scenarios with varying network conditions. Each scenario represents a combination of regular (R) and weak (W) network signal strength over five user-end devices (S1-S5) and 1 edge device (E). The experimental scenarios are summarized in Table \ref{table:EE}. The regular network has no transmission delay, while we add $20$ms delay to all outgoing packets to emulate the weak connection behavior. Each experimental scenario in Table \ref{table:EE} shows the network condition of the specific device. Putting together the five different user devices and one edge device forms a unique combination of varying network conditions per each experimental scenario.

\begin{table}[t]
\small
\centering
\caption{Experiment Environment Setup. $R$ and $W$ represent \textit{Regular} and \textit{Weak} network condition, respectively.}
  \begin{tabular}{c c  c c c c c}
    \toprule
     \parbox[c]{2.7cm}{\hrule height 0pt width 0pt \centering \textbf{Experiment}}&
     \parbox[c]{0.5cm}{\hrule height 0pt width 0pt \centering \textbf{S1}}&
     \parbox[c]{0.5cm}{\hrule height 0pt width 0pt \centering \textbf{S2}}&
     \parbox[c]{0.5cm}{\hrule height 0pt width 0pt \centering \textbf{S3}}&
     \parbox[c]{0.5cm}{\hrule height 0pt width 0pt \centering \textbf{S4}}&
     \parbox[c]{0.5cm}{\hrule height 0pt width 0pt \centering \textbf{S5}}&
     \parbox[c]{0.5cm}{\hrule height 0pt width 0pt \centering \textbf{E}}\\
    \midrule
    \parbox[c]{2.7cm}{\centering EXP-A}& R & R & R & R & R & R\\   
    \parbox[c]{2.7cm}{\centering EXP-B}& R & W & R & W & R & W\\   
    \parbox[c]{2.7cm}{\centering EXP-C}& W & W & W & R & R & R\\   
    \parbox[c]{2.7cm}{\centering EXP-D}& W & W & W & W & W & W\\   

    \bottomrule
  \end{tabular}
  \label{table:EE}
\end{table}


\subsection{Experimental Setup}
 The platform consists of five AWS a1.medium instances with single ARM-core as end-devices connected to an AWS a1.large instance as edge device and an AWS a1.xlarge instance as cloud node. 
 Table \ref{table:DS} summarizes device specifications in details. 
 DL model inferences are executed on processor cores on all nodes using ARM-NN SDK \cite{ltd.}. 
 The inference engine is a set of open-source Linux software tools that enables machine learning workloads on ARM-core-based devices. 
 The framework's message passing protocol is implemented using web services deployed at each node. 
 Section 7.2 provides our analysis on framework's setup overhead.

\begin{table}[b!]
\small
\caption{Device Specification}
\centering
  \begin{tabular}{c c  c c c}
    \toprule
     \parbox[c]{1.5cm}{\hrule height 0pt width 0pt \centering \textbf{Node Type}}&
     \parbox[c]{0.7cm}{\hrule height 0pt width 0pt \centering \textbf{vCPUs}}&
     \parbox[c]{1.5cm}{\hrule height 0pt width 0pt \centering \textbf{Memory (GiB)}}&
     \parbox[c]{1.5cm}{\hrule height 0pt width 0pt \centering \textbf{Frequency (GHz)}}&
    \parbox[c]{3cm}{\hrule height 0pt width 0pt \centering \textbf{Architecture}}\\
    \midrule
    End & 1 & 2 & 2.3 &  aarch64\\   
    Edge & 2 & 4 & 2.3 &  aarch64\\   
    Cloud & 4 & 8 & 2.3 &  aarch64\\   
    \bottomrule
  \end{tabular}
  \label{table:DS}
\end{table}

\subsection{Hyper-parameters and RL Training}
An RL agent has a number of hyper-parameters that impact its effectiveness (e.g., learning rate, epsilon, discount factor, and decay rate). 
The ideal values of parameters depend on the problem complexity, which in our case scales with the number of users (i.e., active end-node devices). In order to determine the learning rate and discount factor, we evaluated values between 0 and 1 for each hyper-parameters. We observed that a higher learning rate converges faster to the optimal, meaning the more the reward is reﬂected to the Q-values, better the agent works. We also observed that a lower discount factor is better. This means that the consecutive actions have a weak relationship, so that giving less weight to the rewards in the near future improves the convergence time. Table \ref{table:HV} shows the different problem configurations we used to determine the hyper-parameters. We train the agent with two different learning algorithms (See Section 4.2). Our Q-Learning agent initializes a Q-table with Q-values of zero, and chooses actions using an $\epsilon-greedy$ policy where $\epsilon$ is the exploration rate. 
 We initially set $\epsilon=1$, meaning the agent selects a random action with probability $1$, otherwise it selects an action that gives the maximum future reward (i.e., Q-value) in the given state. Although we perform probabilistic exploration continuously, we decay the exploration by epsilon decay parameter (See Table \ref{table:HV}) per agent invocation. The Deep Q-Learning agent uses different neural network structure for different number of users as the problem complexity changes. We train DNN models with two fully connected layers where the hidden layers have $48$, $64$, $128$ neurons for three, four, and five devices, respectively. We implement the experience replay as a FIFO buffer with size equal to $1000$. In order to update the network, at each step, we randomly sample $64$ records from the buffer and then use them as a mini-batch. We use  $\epsilon-greedy$ policy to train the Deep Q-network, where we initially set the $\epsilon$ equal to $1$.

\begin{table}[t!]
\small
\caption{Hyper-parameter values}
\centering
\resizebox{.7\textwidth}{!}{
  \begin{tabular}{c c  c   c  c }
    \toprule
    &   \multicolumn{2}{c}{\textbf{Q-Learning}} & \multicolumn{2}{c}{\textbf{Deep Q-Learning}} \\     \toprule

     \parbox[c]{1.5cm}{\hrule height 0pt width 0pt \centering \textbf{Number of Users}}&
     \parbox[c]{1.5cm}{\hrule height 0pt width 0pt \centering \textbf{Learning Rate ($\alpha$)}}&
     \parbox[c]{1.5cm}{\hrule height 0pt width 0pt \centering \textbf{Epsilon Decay}}&
    \parbox[c]{1.5cm}{\hrule height 0pt width 0pt \centering \textbf{Learning Rate ($\alpha$)}}&
     \parbox[c]{1.5cm}{\hrule height 0pt width 0pt \centering \textbf{Epsilon Decay}} \\
    \midrule
    1 & 0.9 &  1e-1  & $-$ & $-$  \\   
    2 & 0.9 &  1e-2  & $-$ & $-$  \\   
    3 & 0.9 &  1e-2  & 1e-3 & 0.4\\  
    4 & 0.9 &  1e-3  & 1e-3 & 0.7\\
    5 & 0.9 & 1e-4  & 1e-3 & 0.9\\   
    \bottomrule
  \end{tabular}
  }
  \label{table:HV}
\end{table}

\section{Evaluation Results and Analysis}
In this section, we demonstrate the effectiveness of our online learning based inference orchestration. We evaluate our approach on the multi-layered end-edge-cloud framework, described in Section \ref{sec.online_framework}. Our approach features online reinforcement learning for intelligent orchestration, DL inference services and end-edge-cloud architectures, targeting DL inference performance. ~\cite{sen2019machine} presents state-of-the-art machine learning based orchestration for end-edge-cloud architecture baseline. For a fair comparison, we evaluate our approach against the strategy proposed in ~\cite{sen2019machine}, which integrates the aforementioned features of our approach. 


\subsection{Performance Analysis}

We evaluate our agent's ability to identify the optimal orchestration decision at each invocation. 
Through reinforcement learning, the agent predicts orchestration decisions including offloading policy and DL model configuration to maximize performance and meet the accuracy threshold. At design time, we determine the true optimal configuration in any given conditions of workloads, network, and number of active users using a brute force search. First, we compare our reinforcement learning based Intelligent Orchestrator's (IO) prediction accuracy against this true optimal configuration. Our proposed approach with both \textit{\textbf{Q-Learning}} and \textit{\textbf{Deep Q-Learning}} algorithm has yielded a 100\% prediction accuracy in comparison with the true optimal configuration. Thus, our reinforcement learning based orchestration decisions always converge with the optimal solution. Next, we evaluate our agent's efficacy by comparing it with a representative state-of-the-art~\cite{sen2019machine} baseline in terms of performance and accuracy.
To implement the baseline policy into our framework, we limit the agent to actions that specify offloading decisions $a_{\tau}=\{o^i\}$, using the most accurate DL model. 
We additionally compare fixed orchestrations for points of reference. 
The fixed solution is limited to configurations where all end-devices either (a) perform the most accurate DL inference execution locally, (b) offload to the edge, or (c) offload to the cloud.
In the following subsections, we demonstrate the efficacy of our proposed agent to find the optimal configuration in presence of different number of users (up to five). 
Then, we investigate its ability to adapt to network variations and evaluate its overhead. 
We explain the impact of varying DL models on the performance under different system dynamics and elaborate how the proposed agent follows the defined constraints. 

\subsubsection{User Variability}

To evaluate the user variability, we consider up to five simultaneously active user-end devices, keeping the network constraints constant. We consider five different levels of accuracy thresholds viz., \textit{Min}, \textit{80\%}, \textit{85\%}, \textit{89\%}, and \textit{Max}. 
\textit{Min} refers to the accuracy threshold for computing where no constraint is applied to the learning algorithms (See Equation 4) and \textit{Max} represents the accuracy threshold for computing where the average accuracy constraint is set to $89.9\%$. We present the average response time and average accuracy for each of these thresholds using our proposed approach. For evaluation, we also present the average response time and accuracy metrics achieved with the state-of-the-art baseline approach \cite{sen2019machine}, and three fixed orchestration decisions viz., device only, edge only and cloud only. 
\begin{figure}[t]
\centering
\includegraphics[width=0.97\linewidth]{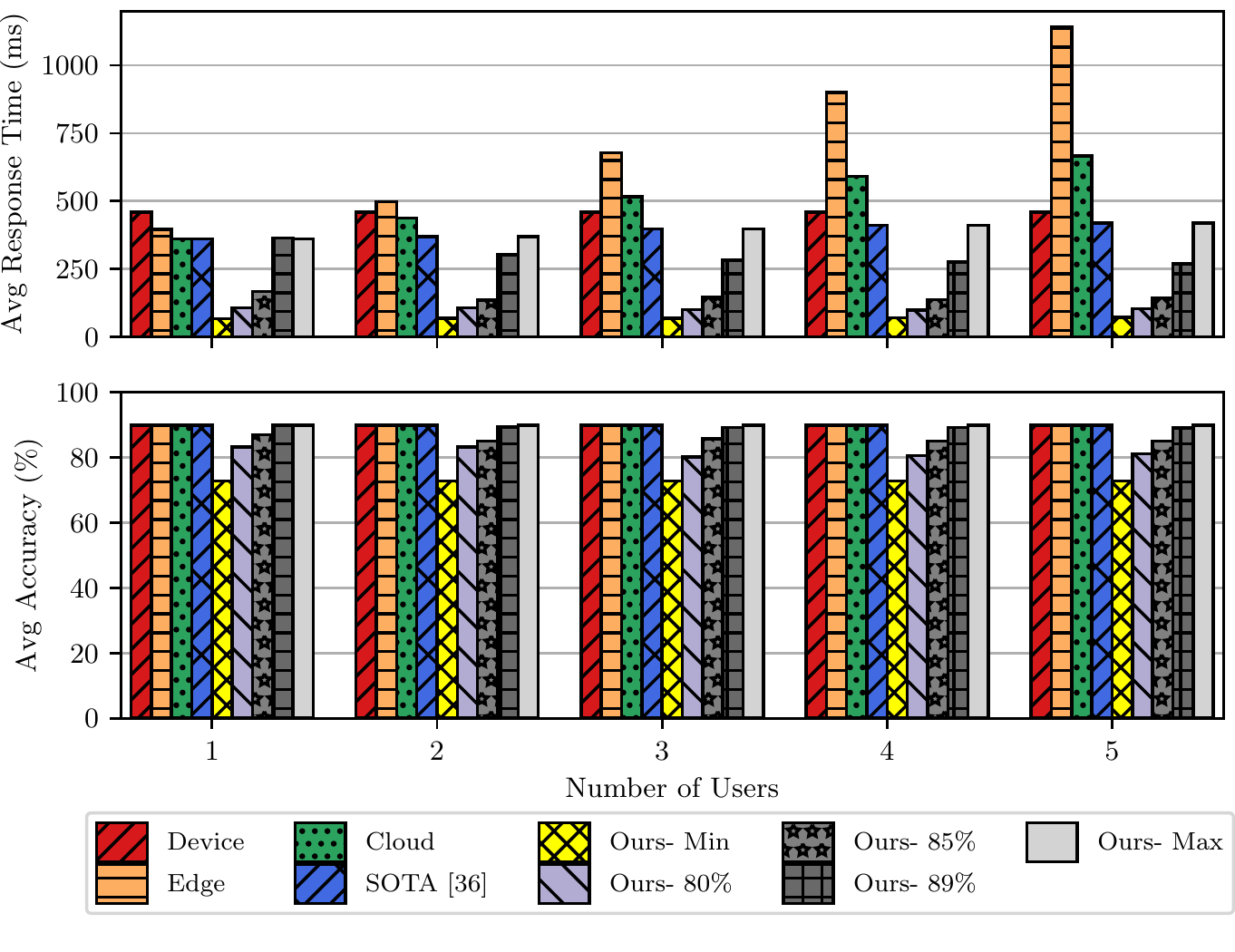}
\caption{Results of the framework within Exp-A for different number of active users.}
\label{fig:perf}
\end{figure}
\newline \textbf{Fixed Strategies} Figure \ref{fig:perf} shows the average response time and accuracy for different numbers of active users for regular network conditions (represented by scenario Exp-A in Table \ref{table:EE}), using different orchestration strategies. 
The x-axis represents the number of active users. 
Each bar represents a different orchestration decision made by using the corresponding orchestration strategy. 
With the device only strategy, each user-end device executes the inference service on the local device. Thus, varying number of users has no effect on the average response time in this case. With the edge and cloud only strategies, simultaneous requests contend for edge and cloud resources. This increases the average response time significantly, as the number of users increase. For instance, the fixed edge only strategy with five active users leads to an average response time of $1140$ms, while it is $665ms$ with cloud only strategy. Higher volume of available resources at the cloud layer results in relatively better average response time in comparison with the edge only strategy. On the other hand, the average response time with the device only strategy is $459ms$, representing the optimal case. 
\newline \textbf{Baseline} With the SOTA \cite{sen2019machine} approach, the average response time remains constant until the number of users is two. This is due to the orchestration decision of distributing the services across edge and cloud layers. As the number of users increase to three, the service requests contend for resources, leading to an increase in the average response time. With the number of users increasing from three through five, the average response time increases, but at a relatively lower rate, exhibiting efficient utilization of the edge and cloud resources. As the number of users increase, the efficiency of the baseline approach over the fixed strategies is more prominent. Both the baseline and fixed strategies are agnostic to model selection and configuration, retaining the maximum prediction accuracy of the inference service. Thus the average accuracy remains constant with the aforementioned strategies, as shown in Figure \ref{fig:perf}.
\newline \textbf{Our proposed solution} Our proposed solution achieves the same average response time in comparison with the baseline for the \textit{Max} accuracy scenario. When the accuracy threshold is relaxed, our reinforcement learning based intelligent orchestrator selects appropriate models (among $d0$-$d7$) to improve the average response time. As the number of users increase, our solution leverages the model selection combined with offloading technique to address the potential increase in response time. With appropriate model selection, our approach reduces the compute intensity, and consequently maintains a lower average response time even with the increasing number of users. Trivially, the average response time with our approach is lower as the accuracy threshold is reduced. However, it should be noted that we enforce the boundaries on tolerable loss of accuracy with our model selection decisions. Figure \ref{fig:perf} shows the average response time and average accuracy with our solution over different scenarios of accuracy thresholds and varying number of users. 
Our solution provides up to $35\%$ improvement in the average response time in comparison with the baseline, within a tolerable loss of $0.9\%$ accuracy. Table \ref{tab:my-table} shows the orchestration decisions of our agent for different numbers of active users, and also over four different experimental scenarios (Table \ref{table:EE}). We present the orchestration decision and the average response time achieved with each decision, for the maximum accuracy threshold scenario. 
\begin{table}[]
\caption{Detailed offloading decisions of our agent for different number of active users in all four experiments (Maximum Accuracy Threshold). For example, in Exp-A, the orchestrator offloads the most accurate DL inference execution ($d0$) to the cloud device ($d0,C$ for end-node $S1$). 
In the presence of five active users, the decisions are $\{d0,E\}$, $\{d0,L\}$, $\{d0,L\}$, $\{d0,C\}$, and $\{d0,L\}$ for end-nodes $S1$ to $S5$, respectively. 
In this case, $S1$,\;$S2$, and $S4$ perform DL inference execution of the $d0$ model locally ($L$). 
$S0$ and $S3$ offload inference execution of the $d0$ model to the edge ($E$) and cloud ($C$), respectively. }
\label{tab:my-table}
\resizebox{.9\textwidth}{!}{
\begin{tabular}{@{}ccccccccc@{}}
\toprule
 &                                 &      & \multicolumn{4}{c}{\textbf{End-node Devices}} & \textbf{} &  \\ \midrule
 &
  \textbf{Experiments} &
  \textbf{Number of Users} &
  \textbf{S1} &
  \textbf{S2} &
  \textbf{S3} &
  \textbf{S4} &
  \textbf{S5} &
  \textbf{Avg Res (ms)} \\ \midrule
\multirow{16}{*}{\rotatebox[origin=c]{90}{\textbf{Decision}}} &
  \multirow{4}{*}{\textbf{Exp-A}} & 1 & $d0,C$ & $-$ & $-$ & $-$ & $-$ & 363.47\\
 &                                 & 2   &   $d0,C$      &   $d0,E$      &    $-$     &    $-$    &   $-$     &   363.17 \\
 &                                 & 3   &   $d0,C$      &   $d0,L$      &    $d0,E$     &   $-$     &   $-$     &   397.53  \\
 &                                 & 4   &   $d0,L$      &   $d0,L$      &    $d0,E$     &   $d0,C$     &   $-$     &   410.35   \\
  &                                 & 5   &   $d0,E$      &   $d0,L$      &   $d0,L$      &  $d0,C$      &   $d0,L$     &  418.91  \\
 \cmidrule(l){2-9} 
 & \multirow{4}{*}{\textbf{Exp-B}} & 1 &   $d0,E$      &   $-$      &   $-$      &  $-$      &  $-$      &   403.30    \\
 &                                 & 2   &   $d0,E$      &   $d0,C$      &   $-$      &   $-$     &   $-$     &   416.78   \\
 &                                 & 3   &   $d0,E$      &    $d0,C$     &   $d0,L$      &  $-$      &  $-$      &   431.90   \\
 &                                 & 4   &   $d0,L$      &   $d0,C$      &   $d0,E$      &   $d0,L$     &  $-$      &   457.96    \\
 &                                 & 5   &   $d0,C$     &  $d0,E$       &   $d0,L$      &  $d0,L$      &  $d0,L$       &   472.88   \\
 \cmidrule(l){2-9} 
 & \multirow{4}{*}{\textbf{Exp-C}} & 1 &   $d0,C$      &   $-$      &  $-$       &   $-$     &  $-$      &   471.65       \\
 &                                 & 2   &  $d0,C$       &   $d0,E$      &  $-$       &  $-$      &   $-$     &   467.80     \\
 &                                 & 3   &   $d0,C$      &  $d0,E$       &   $d0,L$      &  $-$      &  $-$      &   488.21     \\
 &                                 & 4   &   $d0,C$      &  $d0,E$       &  $d0,L$       &  $d0,L$      &   $-$     &   480.70     \\ 
 &                                 & 5   &   $d0,L$       &    $d0,L$      &     $d0,L$     &   $d0,C$      &   $d0,E$   &    464.59   \\ 

 \cmidrule(l){2-9} 
 & \multirow{4}{*}{\textbf{Exp-D}} & 1 &  $d0,L$       &  $-$       &   $-$      &   $-$     &  $-$      &   585.68 \\
 &                                 & 2   &  $d0,E$       &   $d0,C$      &   $-$      &  $-$      &  $-$      &        527.39 \\
 &                                 & 3   &  $d0,L$     &   $d0,C$      &   $d0,E$      &  $-$      &  $-$      &   491.77  \\
 &                                 & 4   &  $d0,L$       &   $d0,C$      &   $d0,E$      &   $d0,L$     &  $-$      &    501.07  \\ 
 &                                 & 5   &   $d0,L$      &   $d0,C$      &   $d0,E$      &   $d0,L$     & $d0,L$       &   506.62   \\ 

 \bottomrule
\end{tabular}}
\end{table}

\begin{table}[]
\caption{Results of the proposed framework for different accuracy constraints for different experiments (five users). For example, in Exp-D with $89\%$ average accuracy constraint, our framework orchestrates $S1$,\;$S2$,\;$S3$, and $S4$ to execute DL inference using model $d4$ locally and offload inference execution using model $d0$ at the cloud. 
However, the baseline obtains the maximum accuracy by executing the most accurate DL inference locally for $S1$,\;$S4$, and \;$S5$ while offloading $d0$ to the edge and cloud for $S3$ and $S2$, respectively.}
\label{tab:networkvariation}
\resizebox{.9\textwidth}{!}{
\begin{tabular}{@{}cccccccccc@{}}
\toprule
 &                                 &      & \multicolumn{5}{c}{\textbf{End-node Devices}} & \textbf{} &  \\ \midrule
 &
  \textbf{Experiments} &
  \textbf{Constraint} &
  \textbf{S1} &
  \textbf{S2} &
  \textbf{S3} &
  \textbf{S4} &
  \textbf{S5} &
  \textbf{Avg Res (ms)} &
  \textbf{Avg Acc (\%)} \\ \midrule
\multirow{16}{*}{\rotatebox[origin=c]{90}{\textbf{Decision}}} &
  \multirow{4}{*}{\textbf{Exp-A}} &
  Min & $d7,L$
   & $d7,L$
   & $d7,L$
   & $d7,L$
   & $d7,L$
   & 72.08 & 72.80 \\
 &                                 & 80\%   &   $d7,L$      &   $d6,L$      &    $d6,L$     &    $d6,L$    &   $d6,L$     &   103.88       & 81.11 \\
 &                                 & 85\%   &   $d2,L$      &   $d6,L$      &    $d5,L$     &   $d6,L$     &   $d5,L$     &   143.81        & 85.06 \\
 &                                 & 89\%   &   $d4,L$      &   $d4,L$      &    $d4,L$     &   $d0,E$     &   $d4,L$     &   269.80        & 89.10 \\
  &                                 & Max   &   $d0,E$      &   $d0,L$      &   $d0,L$      &  $d0,C$      &   $d0,L$     &  418.91         & 89.90 \\
 \cmidrule(l){2-10} 
 & \multirow{4}{*}{\textbf{Exp-B}} & Min &   $d7,L$      &   $d7,L$      &   $d7,L$      &  $d7,L$      &  $d7,L$      &   106.76        & 72.80 \\
 &                                 & 80\%   &   $d6,L$      &   $d3,L$      &   $d6,L$      &   $d6,L$     &   $d6,L$     &   139.92        & 83.23 \\
 &                                 & 85\%   &   $d5,L$      &    $d5,L$     &   $d6,L$      &  $d6,L$      &  $d2,L$      &   176.21        &  85.05\\
 &                                 & 89\%   &   $d4,L$      &   $d4,L$      &   $d0,E$      &   $d4,L$     &  $d4,L$      &   303.50        & 89.10 \\
 &                                 & Max   &   $d0,C$     &  $d0,E$       &   $d0,L$      &  $d0,L$      &  $d0,L$       &   472.88       & 89.90 \\
 \cmidrule(l){2-10} 
 & \multirow{4}{*}{\textbf{Exp-C}} & Min &   $d7,L$      &   $d7,L$      &  $d7,L$       &   $d7,L$     &  $d7,L$      &   119.28        & 72.80 \\
 &                                 & 80\%   &  $d6,L$       &   $d6,L$      &  $d7,L$       &  $d6,L$      &   $d6,L$     &   149.52        & 81.11 \\
 &                                 & 85\%   &   $d5,L$      &  $d6,L$       &   $d5,L$      &  $d6,L$      &  $d5,L$      &   190.76        & 85.47 \\
 &                                 & 89\%   &   $d4,L$      &  $d4,L$       &  $d4,L$       &  $d4,L$      &   $d0,C$     &   318.45        & 89.10 \\ 
 &                                 & Max   &   $d0,L$       &    $d0,L$      &     $d0,L$     &   $d0,C$      &   $d0,E$   &    464.59       & 89.90 \\ 

 \cmidrule(l){2-10} 
 & \multirow{4}{*}{\textbf{Exp-D}} & Min &  $d7,L$       &  $d6,L$       &   $d7,L$      &   $d7,L$     &  $d7,L$      &   158.53        & 72.80 \\
 &     & 80\%   &  $d6,L$       &   $d6,L$      &   $d6,L$      &  $d7,L$      &  $d6,L$      &        182.53   & 81.12 \\
 &                                 & 85\%   &    $d2,L$     &   $d6,L$      &   $d6,L$      &  $d5,L$      &  $d5,L$      &   225.32        & 85.06 \\
 &                                 & 89\%   &  $d4,L$       &   $d4,L$      &   $d4,L$      &   $d4,L$     &  $d0,C$      &    356.75       & 89.10 \\ 
 &                                 & Max   &   $d0,L$      &   $d0,C$      &   $d0,E$      &   $d0,L$     & $d0,L$       &   506.62        & 89.90 \\ 

 \bottomrule
\end{tabular}}
\end{table}

\begin{table}[]
\caption{Results of the state-of-the-art \cite{sen2019machine} in all four experiments.}
\label{tab:networkvariationsota}
\resizebox{.8\textwidth}{!}{
\begin{tabular}{@{}ccccccccc@{}}
\toprule
                                   &       & \multicolumn{5}{c}{\textbf{End-node Devices}} & \textbf{} &  \\ \midrule
 & \textbf{Experiments} & \textbf{S1} & \textbf{S2} & \textbf{S3} & \textbf{S4} & \textbf{S5} & \textbf{Avg Res (ms)} & \textbf{Avg Acc (\%)} \\ \midrule
\multirow{4}{*}{\rotatebox[origin=c]{90}{\textbf{Decision}}} & \textbf{Exp-A} &   $d0,E$      &   $d0,L$      &   $d0,L$      &  $d0,C$      &   $d0,L$     &      418.91     &  89.9\\
                                   & \textbf{Exp-B} &    $d0,C$     &  $d0,E$       &   $d0,L$      &  $d0,L$      &  $d0,L$      &        472.88   &  89.9\\
                                   & \textbf{Exp-C} &   $d0,L$       &    $d0,L$      &     $d0,L$     &   $d0,C$      &   $d0,E$      &        464.59   &  89.9\\
                                   & \textbf{Exp-D} &  $d0,L$        &  $d0,C$        &   $d0,E$      &  $d0,L$       &   $d0,L$     &   506.62        &  89.9\\ \bottomrule
\end{tabular}}
\end{table}

\subsubsection{Network variation}
We consider two possible levels of network connection: (i) a regular network that has low latency, and (ii) a weak network that has high latency. We add $20$ms delay to all outgoing packets to emulate the weak connection behavior. With varying network conditions, there is an increased delay with offloading decisions across the network. Both the baseline and fixed approaches are affected by the weak network conditions, resulting in a higher average response time. The fixed strategies employ the trivial device, edge and cloud only offloading decisions, suffering higher latency. The baseline approach is confined to only an intelligent offloading strategy, which also results in higher average response time inevitably. On the other hand, our proposed solution adapts to varying network conditions by opportunistically exploiting the accuracy trade-offs through model selection. This way, we address for the latency penalty levied by weak network conditions by reducing the compute intensity of the workloads, within the tolerable accuracy bounds. 

Table \ref{tab:networkvariation} shows the orchestration decisions made by our intelligent orchestrator, average response time, and average accuracy achieved over varying networking conditions. 
Each experiment scenario (Exp-A through Exp-D) combines different network conditions for each node in the network (See Table \ref{table:EE}). 
For example, in Exp-A, all the nodes are connected with regular network, whereas in Exp-B, nodes $S1$, $S3$, and $S5$ have regular connections and the rest have weak connections. We set the number of active users to five.  \newline \textbf{Model Selection} Within each experiment scenario, the average response is lower as the accuracy threshold is relaxed. $d0$ through $d7$ represent models with different response time and accuracy levels. For instance, models $d0$, $d4$,\;$d2$,\;$d7$ and $d7$ are selected respectively for accuracy thresholds ranging from \textit{Max} through \textit{Min} in Exp-A. Our proposed orchestrator explores the Pareto-optimal space of model selection and offloading choice, combining the opportunities at application and platform layers simultaneously. For instance in Exp-A, maintaining an accuracy level of 89\% results in an average response time of 269.8ms, by i) setting the models to $d4$,\;$d4$,\;$d4$,\;$d0$, and $d4$ on devices S1-S5, and ii) device configurations to L (local device), L, L, E (edge) and L for S1-S5. However, the average response time can be improved by sacrificing the accuracy within a pre-determined tolerable level. For instance, by lowering the accuracy threshold by 4\% (from 89\% to 85\%), the average response time can be reduced by 88\% (from 269ms to 143ms) by i) setting the models to $d2$,\;$d6$,\;$d5$,\;$d6$, and $d5$ on devices S1-S5, and ii) device configurations to L (local device), L, L, L and L for S1-S5. With varying network conditions, our solution explores the offloading and model selection Pareto-optimal space at run-time to predict the optimal orchestration decisions. 

For example, in Exp-D, our framework obtains $356.75$ms on average response time with significantly weak network connectivity, while it can adapt to regular connectivity in Exp-A to obtain $269.80$ms on average response time. 
In this case, the average accuracy is $89.1\%$ which shows $0.8\%$ error with the maximum average accuracy. 
The baseline ~\cite{sen2019machine} orchestrates the most accurate DL inference execution to obtain $506.62$ms and $418.9$ms average response time in Exp-D and Exp-A, respectively. Orchestration decisions of the baseline approach over different experimental scenarios is summarized in Table \ref{tab:networkvariationsota}.
Although our proposed framework and the baseline can adapt to network variability, our agent provides additional trade-off opportunities to deploy different models combined with offloading technique. 
This leads to up to $35\%$ speedup while sacrificing less than $1\%$ average accuracy. 
      \begin{figure}[t!]
    \centering
    \includegraphics[width=1\linewidth]{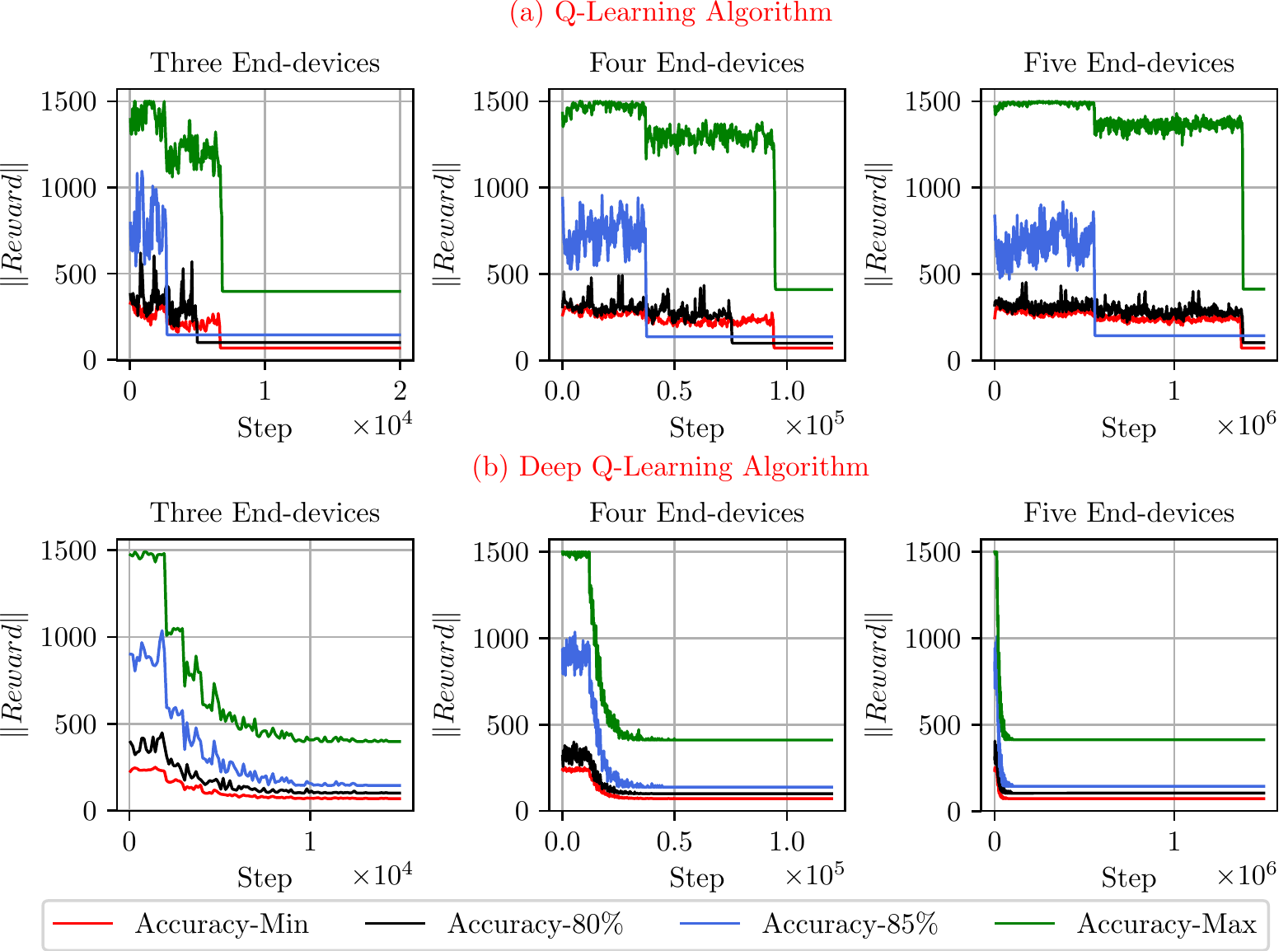}
      \caption{Training overhead for multi-user networks with \textit{\textbf{Q-Learning}} and \textit{\textbf{Deep Q-Learning}} algorithms under different accuracy constraints (See Algorithm 1 and 2, respectively). }
      \label{fig:trainingoverhead}
    \end{figure} 
    
       \begin{figure}[t!]
    \centering
    \includegraphics[width=1\linewidth]{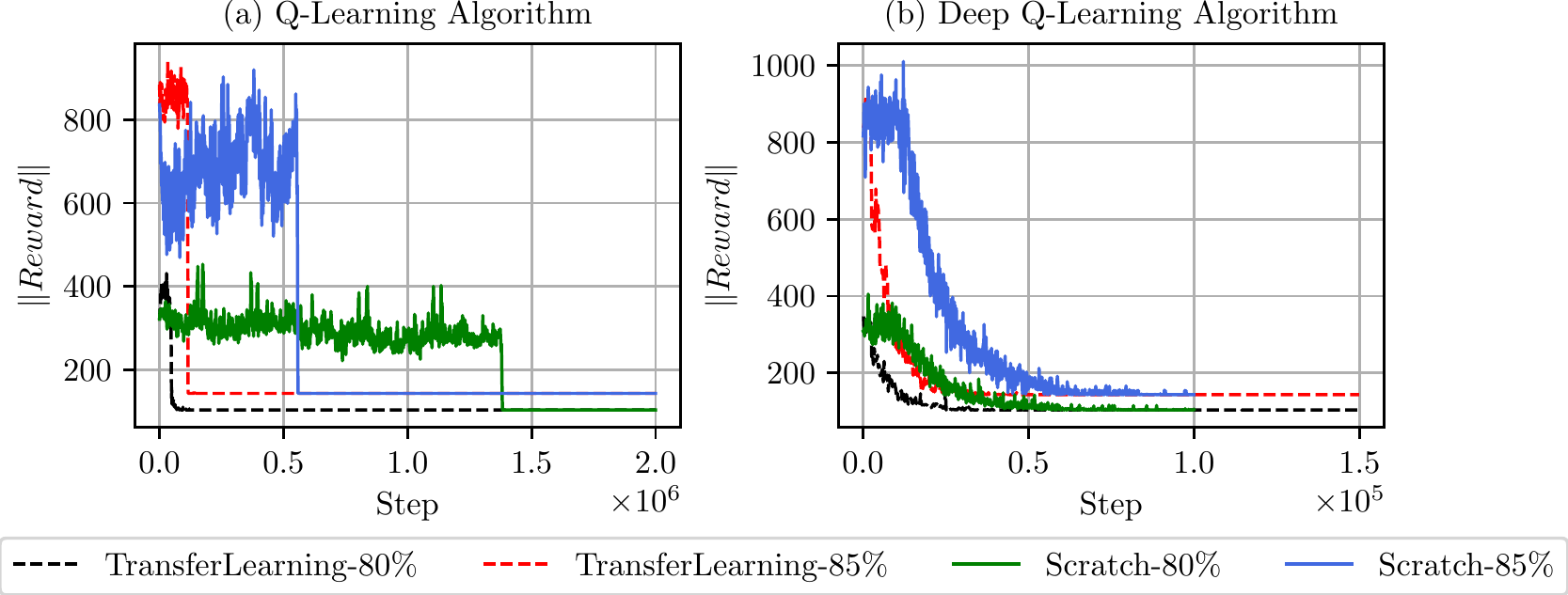}
      \caption{Transfer learning strategy can be used to alleviate the convergence time. In our experiments, the strategy improves the convergence time up to $12.5\times$ and $3.3\times$ for Q-Learning and Deep Q-Learning for five End-devices, respectively. For example, the training phase for Q-Learning algorithm under $80\%$ accuracy constraint converges at $10.5\times10^5$ steps. While, using the transfer learning it converges at  $8.2\times10^4$ steps. }
      \label{fig:trainingoverheadtl}
    \end{figure}  

\subsection{Overhead Analysis}

Developing a global RL agent for optimal runtime orchestration decisions in an end-edge-cloud system incurs overhead to multiple sources.
We evaluate the sources of overhead in both exploration and exploitation phases to demonstrate the feasibility of our proposed solution.
\subsubsection{Exploration Overhead}
  We evaluate the time required by the proposed agent for the training phase to identify an optimal policy. 
  Figure \ref{fig:trainingoverhead} shows the training phase for different numbers of end-devices under different accuracy constraints. We train the agent with \textit{\textbf{Q-Learning}} and \textit{\textbf{ Deep Q-Learning}} algorithms under different accuracy constraints (See Figure \ref{fig:trainingoverhead}.(a) and \ref{fig:trainingoverhead}.(b), respectively).
  The convergence time for five devices with different policies are summarized in Table~\ref{tab:convergencetime}. Q-Learning agent converges faster than Deep Q-Learning agent for the three End-devices scenario. However, increasing the number of End-devices leads to the more complex problem. Deep Q-Learning agent converges up to $17.5\times$ faster than Q-Learning agent for the five End-devices scenario. In other words, Deep Q-Learning algorithm converges faster for high-dimensional space problems. Furthermore, SOTA converges faster since the agent only uses limited actions (3 actions for computation offloading) making a low-dimensional space problem.

  In addition, we observe that the training phase can be accelerated by exploiting previous experiences in similar scenarios known as transfer learning strategy. Figure \ref{fig:trainingoverheadtl} shows that the strategy can alleviate the convergence up to $12.5\times$ and $3.3\times$ for Q-Learning and Deep Q-Learning algorithms, respectively. In the transfer learning strategy, we train a model with minimum accuracy threshold from scratch. Then, we initialize model with the trained model to reduce the convergence time. In conclusion, the Deep Q-Learning algorithm with the transfer learning strategy can speedup the convergence time up to $57.7\times$ in comparison with Q-Learning algorithm for the five End-devices scenario.

  \begin{table}[t!]
\caption{Training convergence time for three, four , and five End-devices with Q-Learning and Deep Q-Learning algorithms compared with SOTA~\cite{sen2019machine} and Bruteforce strategy (See Section 4)}.
\label{tab:convergencetime}
\resizebox{\textwidth}{!}{
\begin{tabular}{@{}cccccc@{}}
\toprule
\textbf{Number of Users} & \textbf{Constraint} & \textbf{Q-Learning } (step \#) & \textbf{Deep Q-Learning} (step \#) & \textbf{SOTA}~\cite{sen2019machine} & \textbf{Bruteforce} (step \#)  \\ 
\midrule
\multirow{4}{*}{\textbf{3}} &Min                 &    $6.6\times10^3$  &   $1.0\times10^4$  & - & $6.6\times10^8$\\
&$80\%$                &     $1.8\times10^3$        &   $1.0\times10^4$    & - & $6.6\times10^8$\\
&$85\%$                &     $0.8\times10^3$        &    $1.0\times10^4$    & - & $6.6\times10^8$\\
&Max                 &     $6.7\times10^3$        &    $1.0\times10^4$    & $2.0\times10^3$ & $6.6\times10^8$ \\
\midrule
\multirow{4}{*}{\textbf{4}} &Min                 &   $9.0\times10^4$          &   $3.0\times10^4$  &- &$5.3\times10^{10}$\\
&$80\% $               &     $8.0\times10^4$                 &    $4.0\times10^4$                  &- &$5.3\times10^{10}$\\
&$85\%$                &     $4.0\times10^4$                &   $4.0\times10^4$                   & - &$5.3\times10^{10}$\\
&Max                 &     $9.0\times10^4$  &         $4.0\times10^4$       & $5.0\times10^3$ &$5.3\times10^{10}$\\
\midrule
\multirow{4}{*}{\textbf{5}} &Min                 &    $10.5\times10^5$        &   $6.0\times10^4$    & - &$4.2\times10^{12}$\\
&$80\%$               &    $10.5\times10^5$  & $6.0\times10^4$ & - & $4.2\times10^{12}$\\
&$85\%$                &    $5.6\times10^5$  &   $7.0\times10^4$        & - & $4.2\times10^{12}$\\
&Max                 &    $10.5\times10^5$     &    $7.0\times10^4$    & $2.5\times10^4$ & $4.2\times10^{12}$\\ \bottomrule
\end{tabular}
}
\end{table}

 \subsubsection{Run-time Overhead}
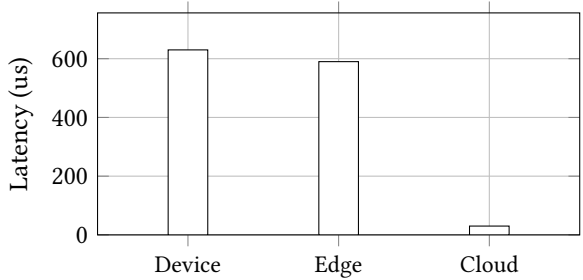
\begin{figure}[h!]
 \centering
  \centering
\begin{tikzpicture}
\begin{axis}[
ybar=2pt,
grid=major,
enlarge x limits={abs=0.6},
ymin=0,
width  = 8cm,
height = 4.5cm,
bar width=15pt,
ylabel={\normalsize Latency (us)},
xticklabel style={rotate=0, font=\small},
xtick = data,
ylabel near ticks,
table/header=false,
every node near coord/.append style={font=\tiny},
table/row sep=\\,
xticklabels from table={
	 Device\\
	 Edge\\
	 Cloud\\
}{[index]0},
legend columns=3,
enlarge y limits={value=.2,upper},
legend style={at={(0.5,.99)},anchor=north, font=\small}
]
\addplot [draw=black, fill=white] table[x expr=\coordindex,y index=0]{630\\590\\30\\};
\pgfplotsinvokeforeach{0,1,2,3}{\coordinate(l#1)at(axis cs:#1,0);}
\end{axis}

\end{tikzpicture}
\caption{Resource Monitoring Overhead}
\label{fig:RSO}
\end{figure}

 The agent is invoked periodically at runtime, imposing overhead on DL inference execution. 
 We evaluate the following components individually:\\
 (a) \textit{Resource Monitoring:} A continuous resource monitoring service imposes runtime overhead in terms of DL inference response time. 
 Figure \ref{fig:RSO} shows that the latency overhead for all layers is negligible (less than $0.8\%$ of minimum response time overall).
\\(b) \textit{Message Broadcasting:} Sharing resource usage and orchestration decision information over the network potentially increases DL inference response time. 
Table \ref{table:SD} shows the additional network latency for different network conditions. 
The \textit{request} is the latency required to send an input image to a higher layer, and dominates the sources of network overhead. 
We observe that the broadcasting, in total, does not impose more than $2\%$ of overall response time.
\\(c) \textit{Intelligent Orchestrator:} The Q-Learning agent's logic itself takes on average $0.6$ms to execute in the cloud. While, the Deep Q-Learning agent's step takes $11ms$ on average to execute using NVIDIA RTX 5000 in cloud.
During exploitation, our trained agent identifies the optimal orchestration decision within five invocations.
We conclude that after an agent is trained, the improvements of $35\%$ in average response time compared to prior art justifies the total overhead of our agent.
 \begin{table}[h]
\small
\caption{Message Broadcasting Overhead}
\centering
  \begin{tabular}{c c c}
    \toprule
     \parbox[c]{3cm}{\hrule height 0pt width 0pt \centering \textbf{}}&
     \parbox[c]{2cm}{\hrule height 0pt width 0pt \centering \textbf{Regular}}&
     \parbox[c]{2cm}{\hrule height 0pt width 0pt \centering \textbf{Weak}}\\
    \midrule
    \parbox[c]{3cm}{\centering Request}& 20 ms & 137 ms \\
    \parbox[c]{3cm}{\centering Update}& 0.4 ms & 2 ms \\
    \parbox[c]{3cm}{\centering Decision}& 1 ms & 2 ms\\
    \midrule
    \parbox[c]{3cm}{\centering \textbf{Total}}& 21.4 ms & 141 ms \\

    \bottomrule
  \end{tabular}
    \vspace{1mm}
  \label{table:overhead}
\end{table}
\section{Conclusion}
Cross-layer optimization that considers both model optimization and computation offloading together provides an opportunity to enhance performance while satisfying accuracy requirements. 
In this paper, for the first time, we proposed an online learning framework for DL inference in end-edge-cloud systems by coordinating tradeoffs synergistically at both the application and system layers. 
The proposed reinforcement learning-based online learning framework adopts model optimization techniques with computation offloading to find the minimum average response time for DL inference services while meeting an accuracy constraint.
Using this method, we observed up to $35\%$ speedup for average response time while sacrificing less than $\%0.9$ accuracy on a real end-edge-cloud system when compared to prior art.
Our approach shows that online learning can be deployed effectively for orchestrating DL inference in end-edge-cloud systems, and opens the door for further research in online learning for this important and growing area. 
\bibliographystyle{ACM-Reference-Format}
\bibliography{ref,related}
\end{document}